\documentclass[journal=acsodf,manuscript=article]{achemso}

\usepackage[version=3]{mhchem}
\usepackage{xcolor}
\usepackage{comment}
\usepackage{hyperref}
\usepackage{algorithm}
\usepackage{algpseudocode}
\usepackage{amsmath}
\usepackage{amssymb}
\usepackage{wasysym}
\usepackage{seqsplit}

\newcommand{\tbf}[1]{\textbf{#1}}
\newcommand{\egc}{$E_{\mathrm{gc}}$}
\newcommand{\egb}{$E_{\mathrm{gb}}$} 
\newcommand{\eea}{$E_{\mathrm{ea}}$} 
\newcommand{\ei}{$E_{\mathrm{i}}$}
\newcommand{\xc}{$X_{\mathrm{c}}$}
\newcommand{\nc}{$n_{\mathrm{c}}$}

\author{Makoto Takamoto}
\affiliation{NEC Laboratories Europe, Heidelberg, Germany}
\email{makoto.takamoto@neclab.eu}
\author{Chaitanya Sampara}
\affiliation{Heidelberg University, Heidelberg, Germany}
\author{Viktor Zaverkin}
\affiliation{NEC Laboratories Europe, Heidelberg, Germany}
\altaffiliation{Present address:
INM -- Leibniz Institute for New Materials, Saarbrücken, Germany;
Saarland University, Department of Computer Science, Saarbrücken, Germany;
German Research Center for Artificial Intelligence (DFKI), Saarbrücken, Germany}
\author{Daniel O\~noro-Rubio}
\affiliation{NEC Laboratories Europe, Heidelberg, Germany}
\author{Mathias Niepert}
\affiliation{NEC Laboratories Europe, Heidelberg, Germany}
\alsoaffiliation{University of Stuttgart, Institute for Artificial Intelligence, Stuttgart, Germany}

\title[Towards Multiscale Molecular Representation Learning through Geometry-Based Fragmentation for Polymer Property Prediction]{Towards Multiscale Molecular Representation Learning through Geometry-Based Fragmentation  for Polymer Property Prediction}

\abbreviations{IR,NMR,UV}
\keywords{American Chemical Society, \LaTeX}

\begin{document}

\begin{abstract}
It is well established that molecular substructures can contribute unequally to chemical and physical properties, with specific motifs or side chains often playing dominant roles. Fragment-based representations have therefore emerged as an effective strategy for incorporating chemically meaningful inductive biases, as exemplified by molecular fingerprints and fragment-based attention mechanisms. However, many existing fragmentation-based machine learning methods rely on predefined chemical rules, which may yield suboptimal fragmentation for previously unseen or chemically unfamiliar structures and often overlook the importance of multiscale molecular representations. 
To address these limitations, we formulate molecular fragmentation 
as a radius-constrained optimization problem inspired by the k-center formulation, where the number of clusters is not fixed a priori but emerges from the specified radius,
and propose a geometry-based fragmentation method that partitions arbitrary molecular or material structures using largely geometric information, without relying on predefined chemical fragmentation rules. 
The method provides flexible control over fragment size, enabling the construction of multiscale fragment representations for scale-adaptive modeling. Building on this framework, we develop a geometric fragment-based attention mechanism that integrates multiscale structural information and incorporate it into a multimodal machine learning model combining p-SMILES representations with three-dimensional (3D) conformational information. Experiments on a benchmark polymer dataset show that incorporating 3D conformational information improves prediction accuracy for several properties over models based exclusively on p-SMILES representations. Explainable artificial intelligence analyses further provide insights into the physical relevance of the learned representations, highlighting the potential of our approach to improve both the accuracy and interpretability of polymer informatics models.
\end{abstract}

\section{Introduction}

Polymer informatics, which leverages data-driven approaches to predict polymer properties, has advanced significantly in recent years, driven by developments in machine learning, high-throughput computation, and the growing availability of materials data. Early efforts, such as the Polymer Genome project \cite{kim2018polymer}, established foundational datasets comprising hundreds of organic polymers with both computed (e.g., bandgaps and dielectric constants) and experimental properties (e.g., glass transition temperatures), enabling the development of multi-task machine learning models. In these initial studies, polymer representations were typically constructed using handcrafted descriptors or fingerprints based on chemical intuition. For instance, neural networks trained on such features demonstrated reasonable predictive performance across multiple properties \cite{kim2018polymer}. However, these approaches rely on expert-designed representations, which can be computationally expensive to generate and may limit their generalizability.

More recently, transformer-based language models have transformed molecular representation learning by directly encoding Simplified Molecular Input Line Entry System (SMILES) strings \cite{weininger1988smiles}, reducing the reliance on manual feature engineering. In polymer informatics, this paradigm is exemplified by models such as PolyBERT \cite{kuenneth2023polybert} and TransPolymer \cite{xu2023transpolymer}. PolyBERT treats canonicalized polymer SMILES (p-SMILES) \cite{ma2020pi1m} as chemical sequences and employs a DeBERTa-based architecture \cite{he2020deberta} \cite{he2020deberta} pretrained on a large collection of synthetically generated polymers via BRICS fragmentation \cite{degen2008art}. Similarly, TransPolymer utilizes a RoBERTa-based encoder \cite{liu2019roberta} with a polymer-specific tokenizer and is pretrained on millions of polymer sequences derived from the PI1M dataset \cite{ma2020pi1m}. These approaches demonstrate the effectiveness of large-scale pretraining for learning data-driven molecular representations. Nevertheless, sequence-based representations primarily encode molecular structures as one-dimensional strings and may not explicitly capture the hierarchical and multiscale organization of molecular structures.

It is well established that the full molecular structure does not contribute equally to chemical and physical properties; rather, specific substructures, such as functional groups and side chains, may play dominant roles. Fragment-based representations have therefore emerged as an effective strategy for incorporating chemically meaningful inductive biases into molecular machine learning. Examples include molecular fingerprints and fragment-based attention mechanisms, which explicitly characterize local and substructural environments \cite{todeschini2008handbook,rogers2010extended,engel2018chemoinformatics,capecchi2020one,chen2021polymer}. Fragment-based approaches, such as FraGAT \cite{zhang2021fragat} for small molecules and PolymerGNN \cite{queen2023polymer} for polymers, further demonstrate the potential of substructural information for improving molecular representation learning. Importantly, molecular properties can depend on structural patterns spanning multiple length scales, motivating the use of hierarchical representations that capture both local and global structural information.

Despite these advances, many existing fragmentation methods rely on predefined chemical rules or domain-specific knowledge. For example, BRICS fragmentation \cite{degen2008art} identifies chemically meaningful bonds according to predefined decomposition rules. While such approaches can provide chemically interpretable fragments, their applicability may be limited when encountering previously unseen, structurally diverse, or chemically unfamiliar molecular and material systems. Moreover, rule-based fragmentation schemes generally provide limited flexibility in controlling fragment size and do not naturally support systematic construction of multiscale representations. These limitations motivate a more general fragmentation strategy that does not depend on predefined chemical knowledge and can be applied consistently across diverse molecular and material structures.

In this work, we address this challenge by introducing a geometry-based fragmentation method that formulates molecular fragmentation as 
a radius-constrained optimization problem inspired by the k-center formulation, where the number of clusters is not fixed a priori but emerges from the specified radius. 
The proposed method selects representative atoms and partitions a molecular or material structure according to geometric proximity, thereby defining fragments using geometric information rather than chemistry-specific fragmentation rules. By varying a pooling scale-length parameter, the characteristic fragment size can be explicitly controlled, enabling the systematic construction of multiscale fragment representations. This geometric formulation provides a general and flexible framework for partitioning arbitrary molecular structures~\footnote{In principle, the proposed formulation is readily applicable to periodic material systems by incorporating periodic boundary conditions.} and offers a natural route toward scale-adaptive molecular representations.

Building on this framework, we develop a geometric fragment-based attention mechanism and incorporate it into a multimodal machine learning model for polymer property prediction. The model combines p-SMILES representations with three-dimensional (3D) conformational information, while the proposed geometric fragmentation mechanism provides multiscale structural information to the 3D representation. In this setting, the multimodal polymer model serves as a showcase application of the proposed fragmentation method rather than as the primary objective of this work. Experiments on a benchmark polymer dataset show that incorporating 3D conformational information improves prediction accuracy for several properties compared with models based exclusively on p-SMILES representations, while the geometric fragment-based branch contributes complementary structural information.

Finally, we employ gradient-based explainable artificial intelligence (XAI) techniques using Captum \cite{kokhlikyan2020captum} to examine the learned representations. The attribution analysis indicates that the model assigns substantial importance to chemically relevant functional groups, including carbonyl-, oxygen-, sulfur-, and nitrogen-containing moieties. These results provide evidence that the learned multimodal representations can capture physically meaningful structural information despite being defined without chemistry-specific fragmentation rules. Overall, this work establishes geometry-based fragmentation as a general, domain-independent approach for constructing controllable multiscale molecular representations and demonstrates its potential through a multimodal application to polymer informatics.

\section{Materials and Methods}

\subsection{Geometric Fragmentation}

\begin{figure}
  \centering
  \includegraphics[trim=7cm 4cm 2cm 6cm, 
                   clip,width=0.47\textwidth]{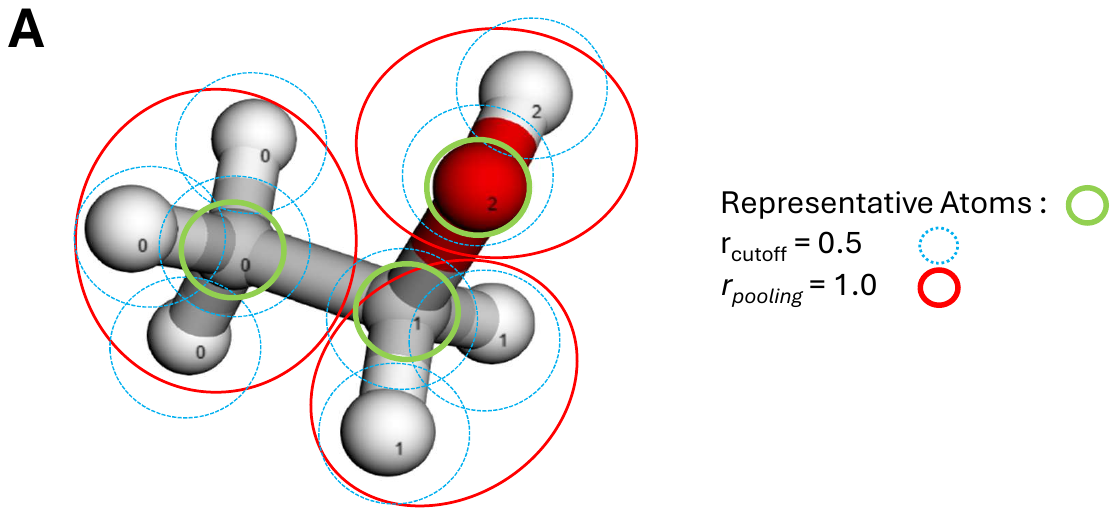}
  \includegraphics[trim=0cm 2cm 0cm 2cm, 
                   clip,width=0.47\textwidth]{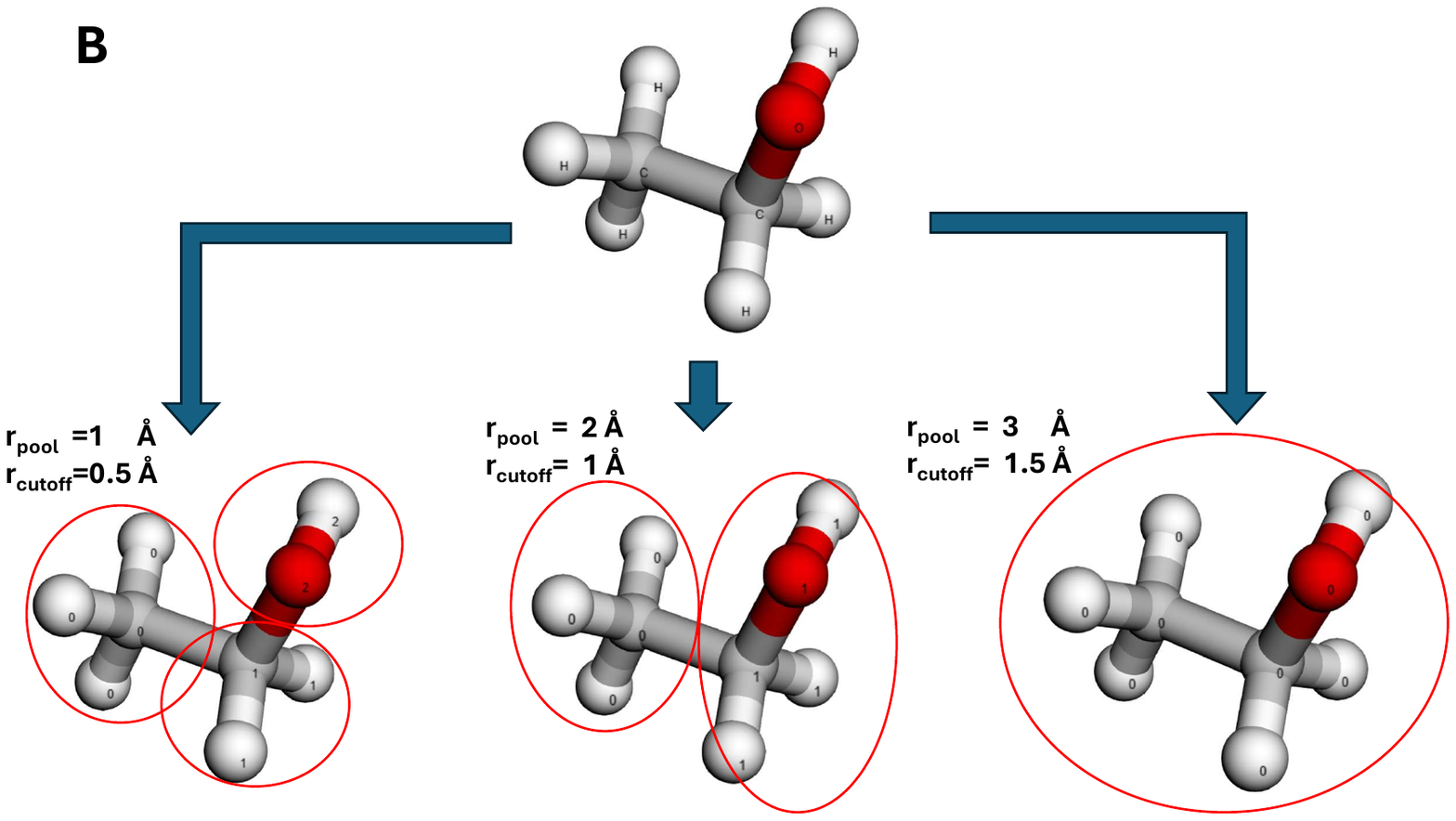}
  \caption{Schematic illustration of the geometric fragment of a monomer, [*]CC(O)[*].
    (A) Example of a fragmentation result with $r_{\rm pool} = 1\,\text{\AA}$ and $r_{\rm cutoff} = 0.5\,\text{\AA}$ with an explicit description of the representative atoms, $r_{\rm pool}$ and $r_{\rm cutoff}$. Here, $r_\mathrm{pool}$ sets the spacing of the representative atoms (centers), whereas $r_\mathrm{cutoff}$ is an atom-centered radius that determines fragment membership. Following the ASE neighbor-list convention, two atoms are neighbors when their radius-$r_\mathrm{cutoff}$ spheres overlap, so the effective pairwise cutoff is $2 \times \mathrm{cutoff}$. (B) Illustration of fragmentation at multiple scales, with $r_{\rm pool} = 1, 2, 3\,\text{\AA}$ and corresponding $r_{\rm cutoff} = 0.5, 1.0, 1.5\,\text{\AA}$, respectively.}
  \label{fig:GF}
\end{figure}

While several molecular fragmentation methods have been proposed, including BRICS, many rely on predefined chemical rules and may therefore produce suboptimal fragmentation schemes when applied to previously unseen or chemically unfamiliar structures. Moreover, these approaches typically provide limited control over the characteristic fragment length scale, which is important for incorporating chemistry-informed inductive biases into molecular representations, as exemplified by recent multiscale molecular fingerprints \cite{rogers2010extended,capecchi2020one}.

In this section, we introduce a geometry-based fragmentation method that partitions arbitrary molecular or material structures using largely geometric information, without requiring chemistry-specific knowledge. This provides an advantage over existing domain-dependent approaches and enables the characteristic fragment size to be flexibly controlled. Consequently, the proposed method facilitates the construction of multiscale fragment representations for hierarchical and scale-adaptive modeling.

Our geometric fragmentation method is based on the $k$-center problem \cite{wolf2011facility,sener2018active,zaverkin2022}, which seeks a set of representative centers from a set of points while minimizing the maximum distance between any point and its nearest center. By treating the atoms of a molecule as points in geometric space, we select representative atoms as centers. In contrast to the classical formulation, we do not fix the number of centers $k$ in advance. Instead, we fix a pooling radius $r_\mathrm{pool}$ and seek a set of centers such that every atom lies within $r_\mathrm{pool}$ of its nearest center. This is the radius-constrained form of the $k$-center problem: writing $d(i,c)$ for the distance between atom $i$ and center $c$,
\begin{equation}
\min_{C} |C|\quad\mathrm{s.t.}\quad\max_{i} \min_{c \in C} d(i,c) \leq r_\mathrm{pool}.
\end{equation}
We approximate its solution with the $k$-Center-Greedy strategy\cite{GONZALEZ1985293}, in which each new center is iteratively selected as
\begin{equation}
c_{n+1} = \arg\max_{i} \min_{c \in C_n} d(i,c),
\end{equation}
thereby selecting the atom that is farthest from its nearest previously selected center. Centers are added until the covering radius $\max_{i} \min_{c \in C} d(i,c)$ falls below $r_\mathrm{pool}$, so that the pooling distance directly controls the number and spatial separation of the representative atoms, which are subsequently used to construct molecular fragments.

\begin{algorithm}[t]
\caption{Geometric Fragmentation}
\label{alg-1}
\begin{algorithmic}[1]
\Require Coordinates $\{R_i\}$, atomic numbers $\{Z_i\}$, pooling distance $r_{\text{pool}}$, cutoff distance $r_{\text{cutoff}}$
\Ensure Disjoint fragments $\{F_1,\dots,F_m\}$ covering all atoms
\State $\mathcal{H} \gets \{ i : Z_i > 1 \}$ \Comment{centers restricted to heavy atoms}
\State Choose an origin $O$;\ \ $c_0 \gets \arg\max_{i \in \mathcal{H}} \|R_i - O\|^2$;\ \ $\mathcal{S} \gets \{c_0\}$
\For{each $i \in \mathcal{H}$}
    \State $D_i \gets \|R_i - R_{c_0}\|^2$ \Comment{squared distance to nearest center}
\EndFor
\While{$\max_{i \in \mathcal{H}} D_i \ge r_{\text{pool}}^2$} \Comment{Stage 1: select centers via $r_{\text{pool}}$}
    \State $j \gets \arg\max_{i \in \mathcal{H}} D_i$;\ \ $\mathcal{S} \gets \mathcal{S} \cup \{j\}$
    \For{each $i \in \mathcal{H}$}
        \State $D_i \gets \min(D_i,\ \|R_i - R_j\|^2)$ \Comment{update distances}
    \EndFor
\EndWhile
\For{each center $s \in \mathcal{S}$} \Comment{Stage 2: build fragments via $r_{\text{cutoff}}$}
    \State $F_s \gets \{s\} \cup \{ j \neq s : \|R_s - R_j\| \le 2\,r_{\text{cutoff}} \}$
\EndFor
\For{each atom $i$ with $\lvert \mathcal{S}_i \rvert > 1$, where $\mathcal{S}_i = \{ s \in \mathcal{S} : i \in F_s \}$}
    \State $s^\ast \gets \arg\min_{s \in \mathcal{S}_i} \|R_i - R_s\|^2$
    \State $F_s \gets F_s \setminus \{i\}$ \; $\forall\, s \in \mathcal{S}_i \setminus \{s^\ast\}$
\EndFor
\For{each atom $i \notin \bigcup_{s \in \mathcal{S}} F_s$}
    \State $\mathcal{F} \gets \mathcal{F} \cup \big\{ \{i\} \big\}$ \Comment{one singleton per unassigned atom}
\EndFor
\State \Return $\mathcal{F}$
\end{algorithmic}
\end{algorithm}

To adapt this construction to geometric fragmentation, we introduce two characteristic length scales that play distinct roles:
\begin{itemize}
\item A pooling distance $r_{\rm pool}$, which sets the spacing between representative atoms (centers) and thereby the number and characteristic size of the fragments. It enters only the center-selection stage, as the stopping criterion of the greedy farthest-point search.
\item A cutoff distance $r_{\rm cutoff}$, which defines, in the fragment-assignment stage, the atom-centered neighborhood used to populate each fragment around its center.
\end{itemize}

We adopt the atom-centered convention of standard neighbor-list implementations (e.g., ASE\cite{Hjorth2017}): each atom carries a sphere of radius $r_\mathrm{cutoff}$, and two atoms are counted as neighbors when their spheres overlap. The effective pairwise cutoff is therefore $2 \times r_\mathrm{cutoff}$, and with the choice $r_\mathrm{cutoff} = r_\mathrm{pool}/2$ used throughout this work, the effective fragment radius $2 \times r_\mathrm{cutoff}$ coincides with the center spacing $r_\mathrm{pool}$.

The pooling distance $r_{\rm pool}$ serves as the stopping criterion for the greedy center-selection procedure: the iteration terminates when the squared distance of the farthest atom from its nearest selected center falls below $r_{\rm pool}^2$. The cutoff distance $r_{\rm cutoff}$ then governs fragment membership: each representative atom is grouped with the atoms within its atom-centered neighborhood, atoms shared between neighborhoods are assigned to the nearest representative, and any atom lying beyond $2 \times r_\mathrm{cutoff}$ of every representative forms its own singleton fragment. An algorithmic description of the proposed method is provided in Algorithm~\autoref{alg-1}, while a step-by-step example is presented in the Supplemental Material. By varying $r_{\rm pool}$, fragments with different characteristic geometric length scales can be generated systematically (see also \autoref{fig:GF}). During the center-selection procedure, hydrogen atoms are excluded from consideration as representative centers because the primary molecular structure is generally defined by heavier atoms, such as carbon.

\subsection{Geometric Fragment-Based Attention Module}

\begin{figure}
  \centering
  \includegraphics[trim=0cm 2cm 0cm 3cm,  
                   clip,width=0.95\textwidth]{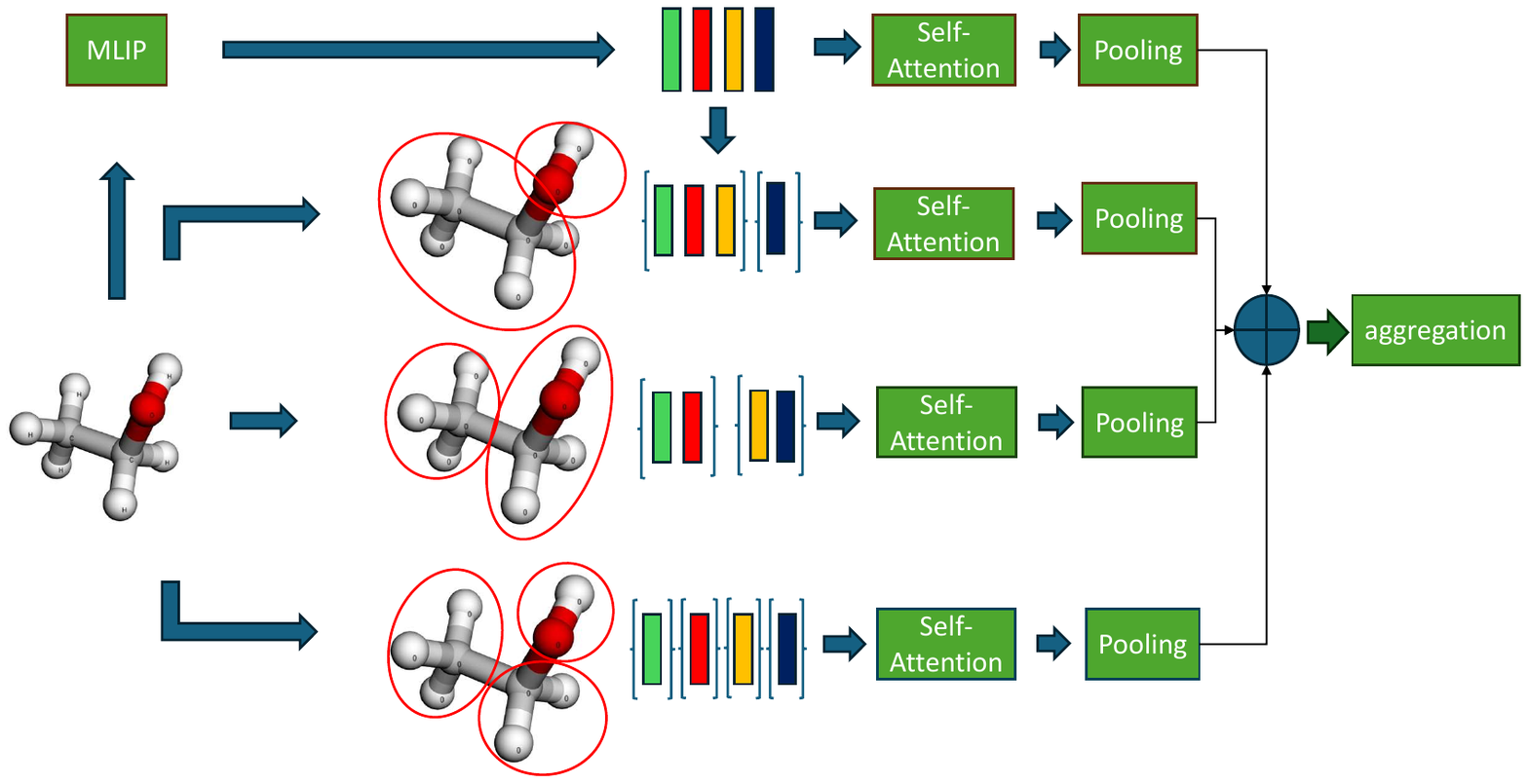}
  \caption{A schematic illustration of the fragment-based attention architecture. The top pathway represents the global information flow, while the lower three pathways correspond to large-, medium-, and small-scale flows, respectively.}
  \label{fig:GFA-structure}
\end{figure}

It is well established that the full molecular structure does not always contribute equally to a chemical reaction; rather, specific substructures, such as side chains, may play a dominant role. Motivated by this observation, fragment-based representations have emerged as an effective strategy for incorporating chemically meaningful inductive biases, leading to approaches such as molecular fingerprints and fragment-based attention mechanisms \cite{zhang2021fragat,wollschlager2024expressivity}. However, many recent fragmentation-based machine learning methods primarily focus on local structural fragments and may overlook the importance of multiscale molecular representations;
Although molecular fingerprints such as ECFP \cite{rogers2010extended} and MAP4 \cite{capecchi2020one} partially capture structural information at different neighborhood scales, they are not specifically designed to construct flexible multiscale molecular representations. Building on this paradigm, we propose a geometric fragment-based attention mechanism that incorporates multiscale structural information while leveraging the scale flexibility of our proposed geometric fragmentation approach, inspired by recent advances in geometry-aware molecular fingerprints \cite{probst2018probabilistic,capecchi2020one}.

In the following, we describe the operation of the proposed geometric fragment-based attention module (see \autoref{fig:GFA-structure}). First, the proposed geometric fragmentation is applied to the 3D molecular conformations and atom-level embedding vectors are assigned to each fragment. In our framework, these embeddings are computed using a machine-learned interatomic potential (MLIP) model, that is, MACE-MP \cite{batatia2023foundation}. To capture multiscale effects \cite{capecchi2020one}, we employ three fragmentation radii, $r_{\rm pool} = 1, 2, 3$~\AA ~to resolve local atomic composition, localized functional groups, and extended topological frameworks, respectively~\footnote{We emphasize that both the selection and the number of fragment radii can be flexibly adjusted according to the target system.}. For each scale, the fragment-level node embeddings are processed by a self-attention module, followed by a pooling operation that produces a single token per scale. As described in \autoref{fig:GFA-structure}, we also retain all the node embeddings to account for global structural information. Finally, these scale-specific tokens are averaged and aggregated into a one-dimensional representation via 1D global depthwise convolution, which is passed to the decoder of the fragmentation branch for downstream tasks.

\subsection{Multimodal Embedding of Polymer Chemistry}

\begin{figure}
  \centering
  \includegraphics[trim=0cm 3cm 0cm 2.5cm, 
                   clip,width=0.95\textwidth]{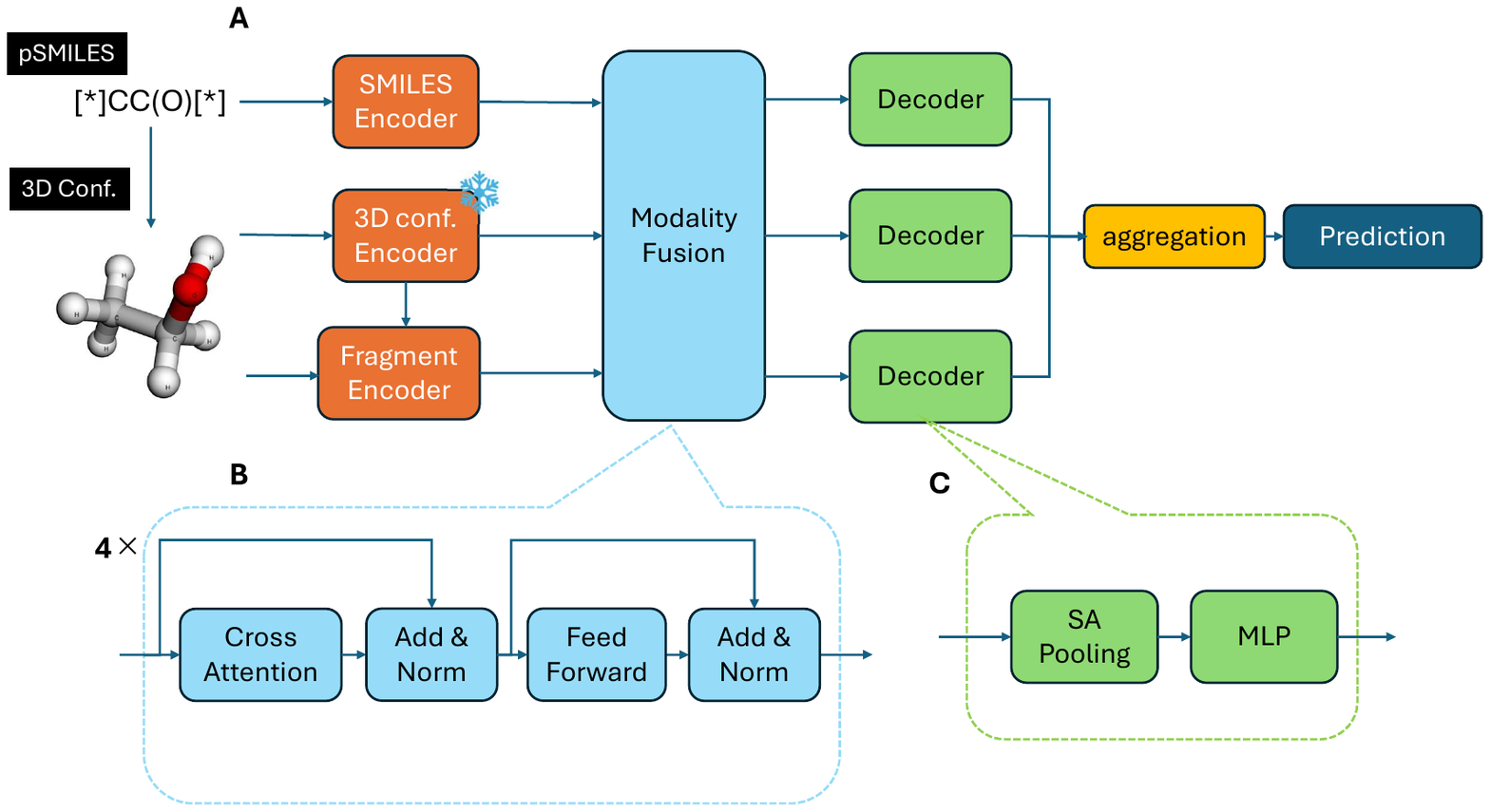}
  \caption{Overview of the multi-modal machine learning model architecture.
    (A) Schematic illustration of the overall model structure. The snow mark above 3D conf. Encoder denotes the model weight of the 3D conformation embedding (MACE-MP) is frozen. 
    (B) Schematic illustration of the modality fusion module, where “Norm” denotes layer normalization and “Add” indicates a residual connection.
    (C) Schematic illustration of the decoder module, where “SA Pooling” refers to self-attention pooling and “MLP” denotes a multilayer perceptron.}
  \label{fgr:Model-structure}
\end{figure}

Although the proposed geometric fragmentation method operates solely on geometric information, its application to molecular structures requires access to three-dimensional (3D) conformations. This makes the method particularly suitable for integration with 3D structural encoders, such as machine-learning interatomic potentials (MLIPs). In contrast, many polymer property prediction datasets provide polymer structures primarily in the form of p-SMILES. Moreover, recent SMILES-based foundation models, such as TransPolymer and PolyBERT, have established strong baselines for polymer property prediction. Building on the proposed geometric fragmentation method, we develop a multimodal model that integrates p-SMILES representations with 3D conformational information. The architecture combines a p-SMILES encoder and a 3D molecular conformation encoder, followed by a modality-fusion module that integrates the complementary information from the two representations (see also \autoref{fgr:Model-structure}). To leverage established models for each modality, we employ PolyBERT \cite{kuenneth2023polybert} and TransPolymer \cite{xu2023transpolymer} as p-SMILES encoders~\footnote{We used the polyBERT training scripts obtained in 2024 to independently train a tokenizer and model from scratch, followed by fine-tuning on the polymer-property data. No pretrained polyBERT weights or tokenizer were used; full training and licensing details are provided in the Supplemental Materials.} and MACE-MP \cite{batatia2022mace,batatia2023foundation} as the 3D conformation encoder. The atom-level representations obtained from MACE-MP are subsequently processed by a fragment encoder based on the proposed geometry-based fragmentation scheme, enabling the incorporation of multiscale structural information into the model.

To demonstrate the applicability of the proposed approach to polymer datasets containing only p-SMILES, we generate 3D molecular conformations from p-SMILES using RDKit and provide the resulting structures as additional inputs alongside the original p-SMILES representations. A detailed description of the procedure for generating 3D conformations from p-SMILES is provided in the Supplemental Material. This multimodal architecture serves primarily as a testbed for demonstrating the utility of the proposed geometric fragmentation method, while also enabling an assessment of the complementary predictive value of 3D conformational information beyond p-SMILES-based representations. 
Note that the weight parameters of the 3D conformation encoder, MACE-MP, are frozen during training to preserve its physically grounded representation of quantum-mechanical interactions, which is essential for accurately capturing the chemical contributions of individual fragments and their impact on the target properties.

\section{Results}

\subsection{Multiscale Analysis of Geometric Fragmentation}

\begin{figure}
  \centering
  \includegraphics[trim=0cm 2cm 0cm 2cm, 
                   clip,width=0.95\textwidth]{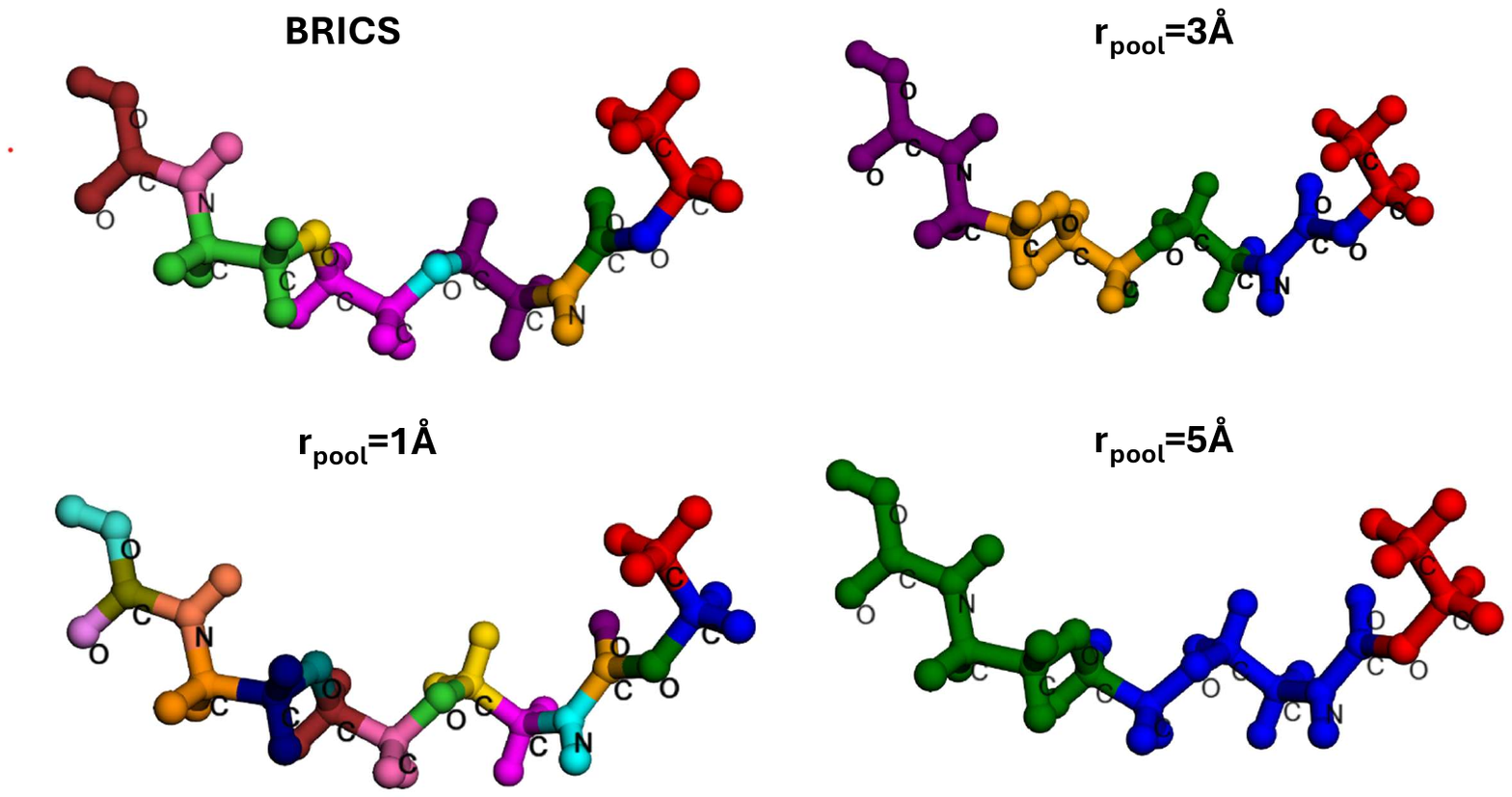}  
  \caption{Fragmentation results of a monomer: [*]CCOC(=O)NCCOCCOCCNC(=O)O[*].} 
  \label{fig:frag-example}
\end{figure}

\autoref{fig:frag-example} illustrates the fragmentation results obtained using BRICS and the proposed geometric fragmentation method with $r_{\rm pool} = 1$, $3$, and $5$ [\AA] for the monomer: 
\seqsplit{[*]CCOC(=O)NCCOCCOCCNC(=O)O[*]}. For the geometric fragmentation method, the cutoff length $r_{\rm cutoff}$ is set to half of the pooling length. While BRICS produces relatively small fragments as a consequence of its rule-based decomposition, the proposed geometric fragmentation method generates fragments whose spatial extent is explicitly governed by the prescribed length scale $r_{\rm pool}$. This distinction highlights a key advantage of the proposed approach: the characteristic spatial scale of the resulting fragments can be directly controlled without relying on predefined chemical decomposition rules.

As is well-known \cite{roberts1965basic,mcmurry2005organic}, molecular structures exhibit characteristic features across a broad range of spatial scales. Individual covalent bonds typically have lengths on the order of $1$--$2$ \AA, whereas chemically relevant local motifs and functional groups extend over several \AA. Longer side chains can span length scales approaching or exceeding $1$ nm. These observations motivate a multiscale fragmentation strategy capable of capturing structural information across different spatial resolutions. By varying $r_{\rm pool}$, the proposed geometric fragmentation method provides a direct and flexible mechanism for constructing fragment representations at different characteristic length scales.

\subsection{Quantitative analysis of multimodal effect}
\begin{table}[th]
  \caption{Benchmark results on polymer datasets. The performance is measured using root mean-square error (RMSE). TP denotes TransPolymer, and PB denotes PolyBERT. The number written in bold font indicates the best performance for each dataset.}
  \label{tbl:benchmark}
  \begin{center}
  \resizebox{\textwidth}{!}{
  \begin{tabular}{lllllllll}
    \hline
    Model Name  &  Modality &   &     & \multicolumn{3}{c}{Test RMSE} &  \\
                & & \egc (eV) & \egb (eV) & \eea (eV) & \ei (eV) & \xc (\%) & EPS & \nc \\
    \hline
    TP (scratch)\textsuperscript{\emph{a}} 
                           & SMILES & 0.58$^{\pm 0.01}$ & 0.60$^{\pm 0.03}$ & 0.36$^{\pm 0.03}$ & 0.46$^{\pm 0.04}$ & 18.9$^{\pm 0.51}$ & 0.57$^{\pm 0.11}$ & 0.099$^{\pm 0.015}$  \\
    TP (pretrained)\textsuperscript{\emph{a}} 
                           & SMILES & 0.44$^{\pm 0.01}$ & 0.53$^{\pm 0.04}$ & 0.31$^{\pm 0.03}$ & \tbf{0.39}$^{\pm 0.07}$ & 17.5$^{\pm 0.69}$ & 0.54$^{\pm 0.05}$ & 0.098$^{\pm 0.017}$  \\
    \hline    
    PB+MACE & SMILES  & 0.44$^{\pm 0.02}$ & 0.48$^{\pm 0.05}$ & \tbf{0.26}$^{\pm 0.01}$ & 0.39$^{\pm 0.05}$ & 17.1$^{\pm 0.59}$ & 0.49$^{\pm 0.04}$ & 0.087$^{\pm 0.013}$ \\
    PB+MACE & 3D  & 0.48$^{\pm 0.03}$ & 0.50$^{\pm 0.05}$ & 0.29$^{\pm 0.03}$ & 0.40$^{\pm 0.04}$ & 17.7$^{\pm 0.94}$ & 0.49$^{\pm 0.06}$ & 0.085$^{\pm 0.016}$ \\
    PB+MACE & SMILES+3D  & \tbf{0.42}$^{\pm 0.01}$ & 0.46$^{\pm 0.04}$ & 0.27$^{\pm 0.02}$ & 0.39$^{\pm 0.05}$ & 17.1$^{\pm 0.48}$ & \tbf{0.47}$^{\pm 0.05}$ & 0.085$^{\pm 0.012}$ \\
    TP+MACE & SMILES  & 0.43$^{\pm 0.01}$ & 0.47$^{\pm 0.04}$ & 0.28$^{\pm 0.02}$ & 0.39$^{\pm 0.05}$ & 17.3$^{\pm 0.33}$ & 0.51$^{\pm 0.06}$ & 0.090$^{\pm 0.013}$ \\
    TP+MACE & 3D  & 0.47$^{\pm 0.03}$ & 0.51$^{\pm 0.04}$ & 0.30$^{\pm 0.03}$ & 0.40$^{\pm 0.05}$ & 18.2$^{\pm 0.57}$ & 0.48$^{\pm 0.04}$ & 0.087$^{\pm 0.012}$ \\
    TP+MACE & SMILES+3D  & \tbf{0.42}$^{\pm 0.01}$ & \tbf{0.45}$^{\pm 0.05}$ & 0.28$^{\pm 0.02}$ & \tbf{0.38}$^{\pm 0.05}$ & \tbf{16.6}$^{\pm 0.43}$ & 0.50$^{\pm 0.05}$ & \tbf{0.083}$^{\pm 0.013}$ \\
    \hline
  \end{tabular}}
  \end{center}
  \footnotesize{\textsuperscript{\emph{a}} For a fair comparison, we changed the TP model from double precision to single precision and re-performed the training using the same dataset as used for our models.}
\end{table}

In this section, we quantitatively evaluate the multimodal model incorporating the geometric-fragmentation-based attention branch. 

For this purpose, we use a widely adopted benchmark dataset for polymers \cite{kim2018polymer} which provides p-SMILES representations together with a range of polymer properties, including the chain bandgap ($E_{\mathrm{gc}}$), bulk bandgap ($E_{\mathrm{gb}}$), electron affinity ($E_{\mathrm{ea}}$), ionization energy ($E_{\mathrm{i}}$), crystallization tendency ($X_{\mathrm{c}}$), dielectric constant (EPS), and refractive index ($n_{\mathrm{c}}$), which are computed using density functional theory (DFT). As described in the Method section, 3D conformations are generated using RDKit. 

As a baseline, we adopt TransPolymer \cite{xu2023transpolymer}, a state-of-the-art model developed for this dataset. Within our multimodal framework, SMILES-based representations are encoded using either PolyBERT \cite{kuenneth2023polybert} or TransPolymer \cite{xu2023transpolymer}, enabling a direct comparison between different sequence-based encoders. For the 3D structural modality, we employ the MACE-MP model \cite{batatia2023foundation} as a conformational encoder to obtain geometry-aware atomic representations. This setup allows us to systematically assess the individual contributions of each modality, as well as their combined effects, to predictive performance across different polymer properties. Following the evaluation protocol of TransPolymer \cite{xu2023transpolymer}, we use five-fold cross-validation and evaluate model performance based on the average root mean square error (RMSE) on the validation sets. The models are pretrained on the PI1M dataset \cite{ma2020pi1m}, including TransPolymer and polyBERT weights, augmented to five times its original size following the procedure described in \cite{xu2023transpolymer}. Detailed experimental settings are provided in the Supplemental Material. 

The benchmark results are presented in \autoref{tbl:benchmark}. Incorporating 3D conformational information improves predictive performance for several polymer properties, demonstrating the complementary value of geometric information beyond p-SMILES representations. Compared with the baseline, our multimodal model achieves an improvement of approximately 5--10\% in predictive performance for several properties in comparison to the baseline. However, the contribution of the 3D modality is property-dependent; \egc \ and \xc \ of TransPolymer exhibit improved performance when 3D structural information is incorporated in addition to p-SMILES representations, whereas the remaining properties show an improvement in the averaged performance that fall within the observed fluctuations. 
We also note that no clear improvement is observed for $E_{\mathrm{i}}$. This may indicate that the available structural and sequence information provides limited additional predictive benefit for this property within the current dataset, imposing an empirical performance ceiling. 

\begin{table}[t]
  \caption{
  Ablation Study of the branch dependence on polymer datasets. The performance is measured using root mean-square error (RMSE). TP denotes TransPolymer, and PB denotes PolyBERT. The number written in bold font indicates a result showing a relatively notable difference from the other results, although the difference remains within one standard deviation.}
  \label{tbl:ablation}
  \begin{center}
  \resizebox{\textwidth}{!}{
  \begin{tabular}{lllllllll}
    \hline
    Model Name  & Branch &   &     & \multicolumn{3}{c}{Test RMSE} &  \\
                & & \egc (eV) & \egb (eV) & \eea (eV) & \ei (eV) & \xc (\%) & EPS & \nc \\
    \hline
    PB+MACE & pSMILES+3D+Fragment & 0.42$^{\pm 0.01}$ & 0.46$^{\pm 0.04}$ & 0.27$^{\pm 0.02}$ & 0.39$^{\pm 0.05}$ & 17.1$^{\pm 0.48}$ & 0.47$^{\pm 0.05}$ & 0.085$^{\pm 0.012}$ \\
    PB+MACE & 3D+Fragment & 0.42$^{\pm 0.01}$ & 0.48$^{\pm 0.05}$ & 0.28$^{\pm 0.02}$ & 0.39$^{\pm 0.05}$ & 17.1$^{\pm 0.47}$ & 0.50$^{\pm 0.05}$ & 0.084$^{\pm 0.012}$ \\
    PB+MACE &  pSMILES+Fragment & 0.42$^{\pm 0.02}$ & 0.46$^{\pm 0.05}$ & 0.27$^{\pm 0.03}$ & 0.38$^{\pm 0.05}$ & 17.1$^{\pm 0.48}$ & 0.51$^{\pm 0.05}$ & 0.085$^{\pm 0.012}$ \\
    PB+MACE & pSMILES+3D & 0.41$^{\pm 0.02}$ & 0.45$^{\pm 0.04}$ & 0.27$^{\pm 0.02}$ & 0.38$^{\pm 0.05}$ & \tbf{17.5}$^{\pm 0.52}$ & 0.48$^{\pm 0.04}$ & 0.081$^{\pm 0.013}$ \\
    \hline
    TP+MACE & pSMILES+3D+Fragment & 0.42$^{\pm 0.01}$ & 0.45$^{\pm 0.05}$ & 0.28$^{\pm 0.02}$ & 0.38$^{\pm 0.05}$ & 16.6$^{\pm 0.43}$ & 0.50$^{\pm 0.05}$ & 0.083$^{\pm 0.013}$ \\
    TB+MACE & 3D+Fragment  & 0.42$^{\pm 0.01}$ & 0.46$^{\pm 0.04}$ & 0.29$^{\pm 0.02}$ & 0.38$^{\pm 0.05}$ & 17.1$^{\pm 0.51}$ & 0.50$^{\pm 0.06}$ & 0.083$^{\pm 0.012}$ \\
    TB+MACE & pSMILES+Fragment  & 0.42$^{\pm 0.01}$ & 0.45$^{\pm 0.04}$ & 0.27$^{\pm 0.02}$ & 0.38$^{\pm 0.05}$ & 16.9$^{\pm 0.80}$ & 0.50$^{\pm 0.04}$ & 0.085$^{\pm 0.014}$ \\
    TB+MACE & pSMILES+3D & 0.41$^{\pm 0.02}$ & 0.46$^{\pm 0.04}$ & 0.27$^{\pm 0.02}$ & 0.39$^{\pm 0.06}$ & 16.8$^{\pm 0.65}$ & 0.48$^{\pm 0.06}$ & 0.083$^{\pm 0.011}$ \\
    \hline
  \end{tabular}}
  \end{center}
\end{table}

The effect of the different model components also depends on the model architecture and target property. We therefore investigate the contribution of the fragment-based attention branches through an ablation study, in which each branch is individually omitted during training\footnote{The scale-dependence study of the fragment-based attention branch is provided in the Supplemental Materials.}. The results are presented in \autoref{tbl:ablation}. Overall, the model exhibits relatively weak dependence on any individual fragmentation branch, with performance differences generally falling within the corresponding error bars. This suggests that the model can exploit complementary information across multiple structural representations rather than relying strongly on a single fragmentation scale. One notable exception is the PolyBERT model trained on \xc\ without the fragmentation branch, for which a degradation in the averaged predictive performance is observed. This result suggests that fragment-level structural information may play a particularly important role in predicting this property. 

\begin{figure}
  \centering
  \includegraphics[clip,width=0.95\textwidth]{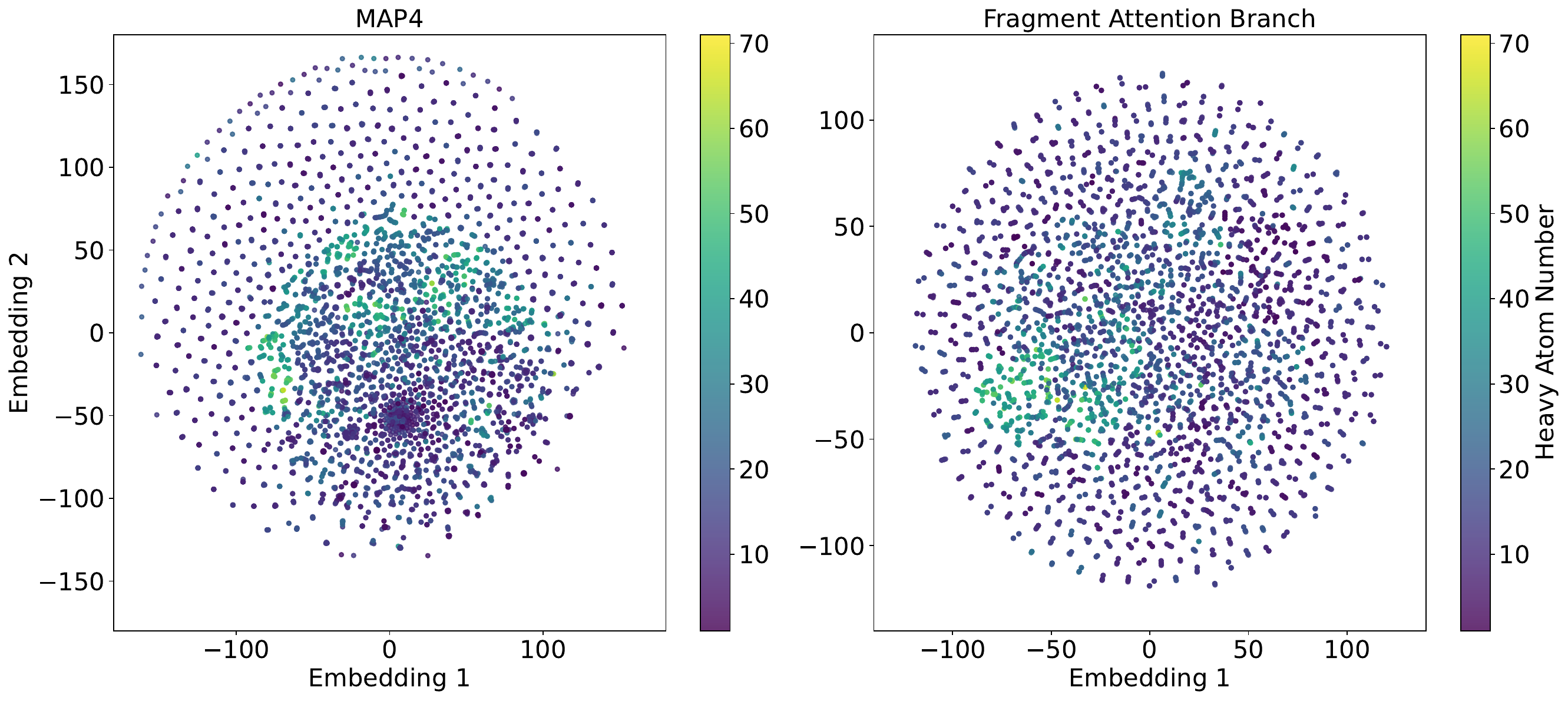}  
  \caption{Scatter plot of latent vector embedded by MAP4 \cite{capecchi2020one} and fragment-attention branch. The embedded vectors are mapped onto two-dimensional space with tSNE. The color map shows the heavy atom number in the monomer.} 
  \label{fig:embed-scatter}
\end{figure}

Finally, we compare the embeddings produced by our geometric-fragmentation-based attention module with MAP4 fingerprints \cite{capecchi2020one}. We use all p-SMILES representations in the polymer dataset \cite{kim2018polymer} for this analysis. For our model, we use the weights fine-tuned on the \egc \ dataset using fold 1. We first compute the MAP4 fingerprints and the embeddings generated by our geometric fragmentation-based attention module, and then apply t-SNE~\cite{van2008visualizing} to project both representations into a two-dimensional space. The results are shown in \autoref{fig:embed-scatter}. The embeddings produced by our fragment attention branch exhibit a relatively uniform and isotropic distribution of the input monomers, whereas the MAP4 fingerprints produce a more non-uniform distribution. The non-uniform distribution of MAP4 fingerprints may be advantageous for rapid molecular screening, which is one of the primary applications of MAP4. In contrast, the more uniform coverage observed in the learned fragment-based representation may provide a potentially useful inductive bias for downstream property prediction by reducing the concentration of representations in specific regions of the molecular space. However, this interpretation should be regarded as a qualitative hypothesis, as the two-dimensional t-SNE projections alone do not establish a direct relationship between representation uniformity and predictive performance. The color map indicates the number of heavy atoms in each monomer. In both representations, we observe a weak clustering effect associated with the number of heavy atoms, suggesting that molecular size remains a relevant factor in the learned representation. 

\subsection{Interpretation of model prediction via Explainable AI method}

\begin{figure}
  \centering
  \includegraphics[trim=0cm 2cm 0cm 2cm, 
                   clip,width=0.95\textwidth]{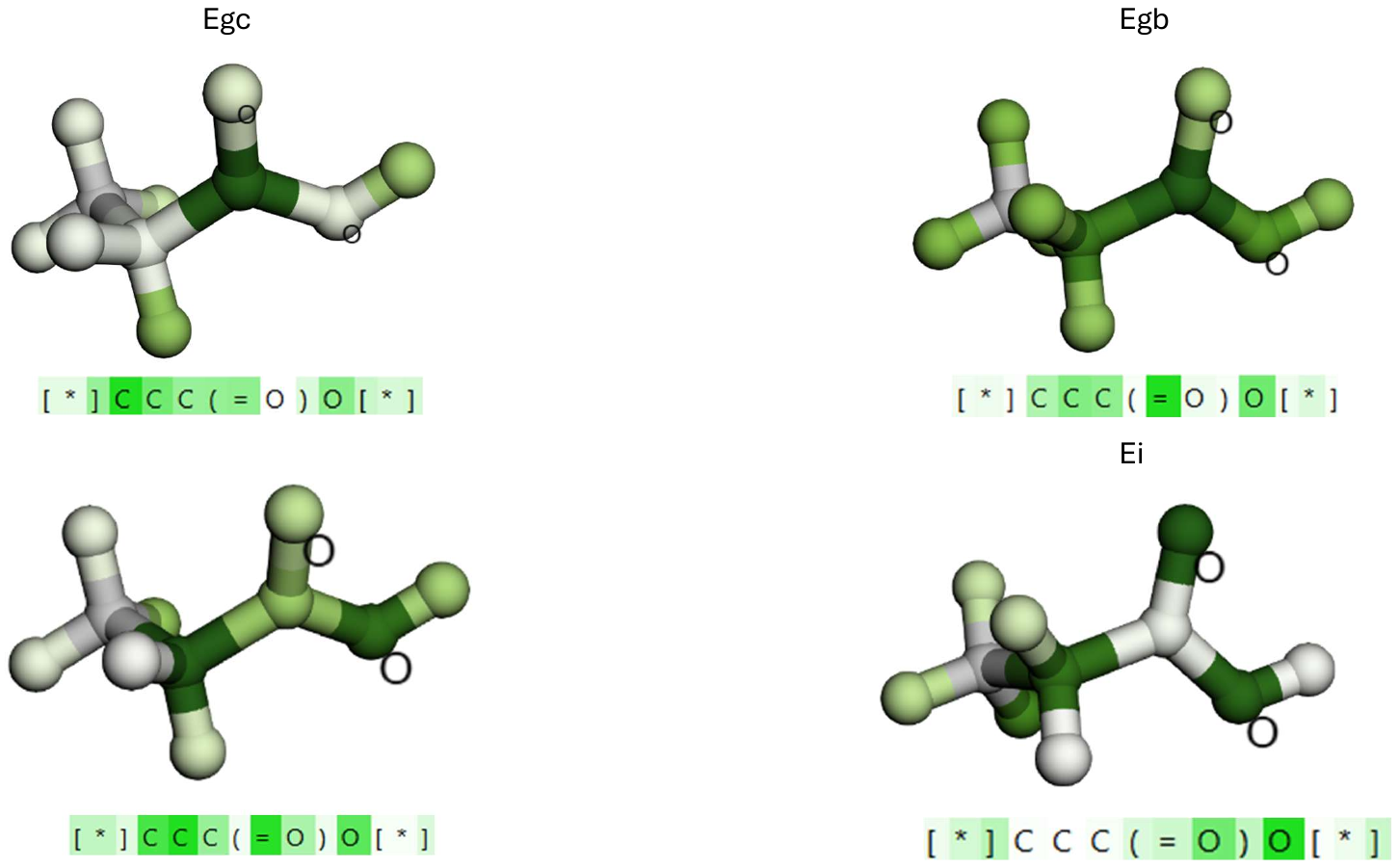}
  \includegraphics[trim=0cm 2cm 0cm 2cm, 
                   clip,width=0.95\textwidth]{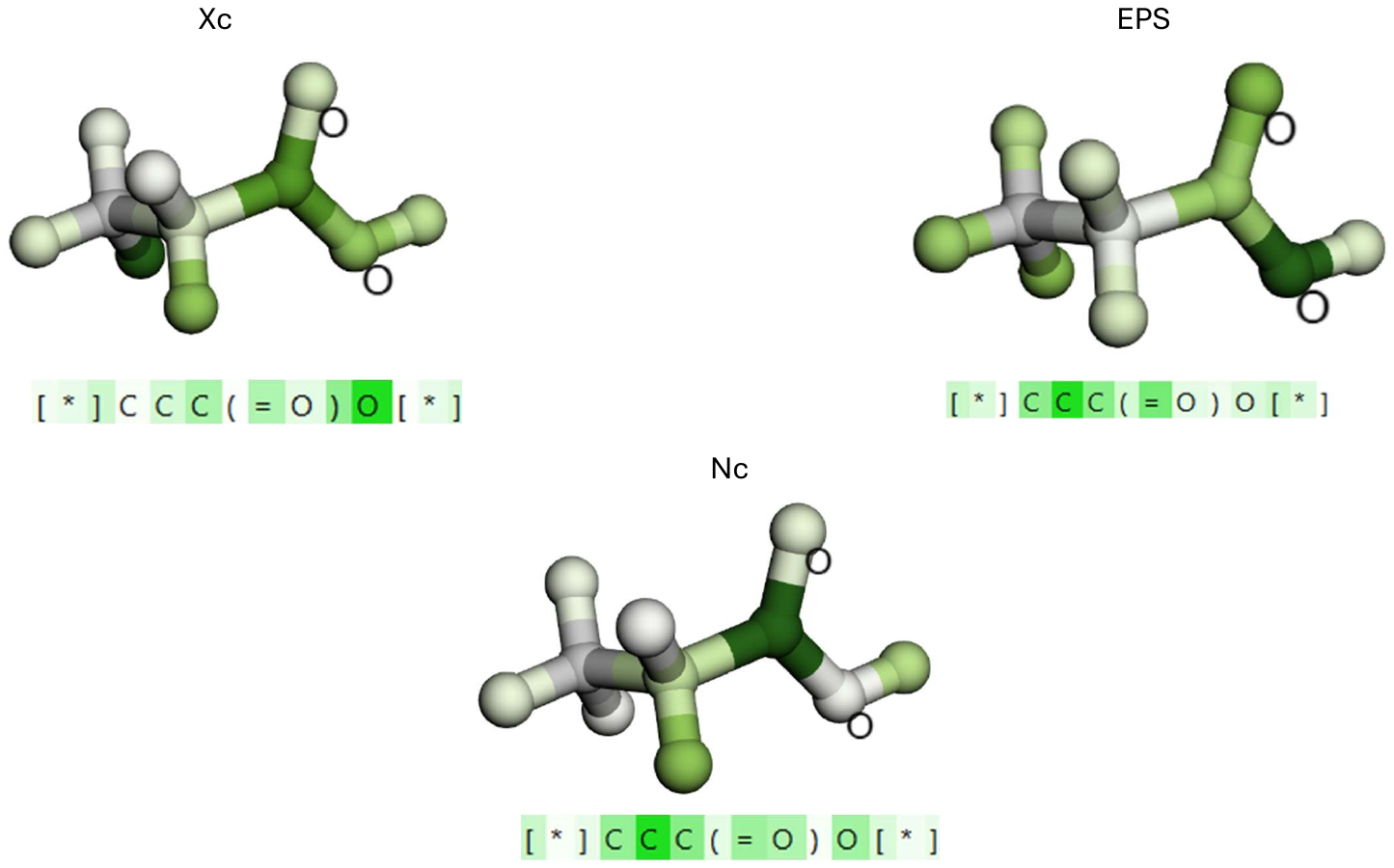}
  \caption{A plot of importance-score of the input structure in the case of [*]CCC(=O)O[*]. 
  Only oxygen atoms are labeled, whereas carbon and hydrogen atoms are left unlabeled for clarity. The plotted importance-score is normalized by the maximum value of the original importance-score.} 
  \label{fig:XAI}
\end{figure}

\begin{figure}
  \centering
  \includegraphics[trim=0cm 2cm 0cm 2cm, 
                   clip,width=0.95\textwidth]{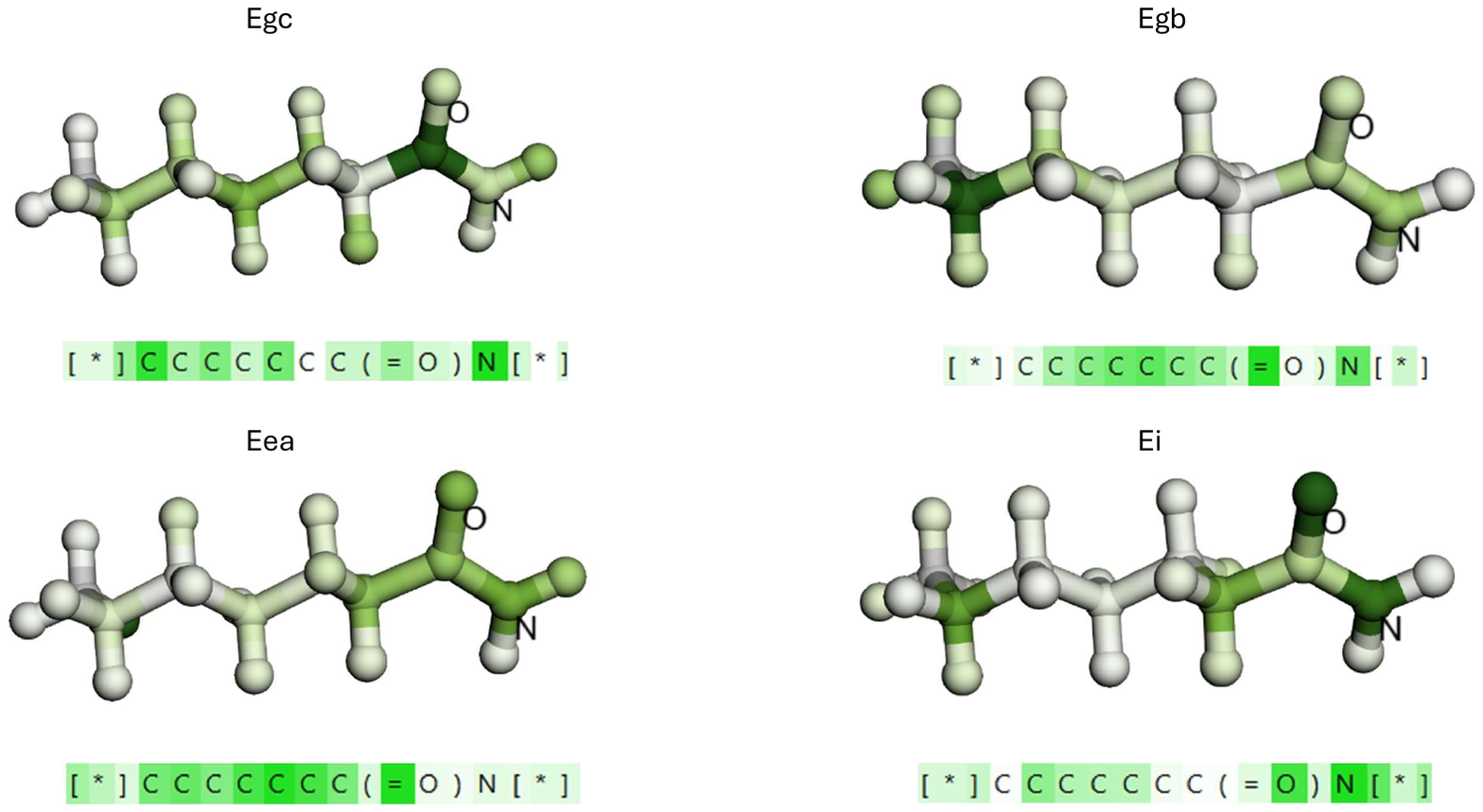}
  \includegraphics[trim=0cm 2cm 0cm 2cm, 
                   clip,width=0.95\textwidth]{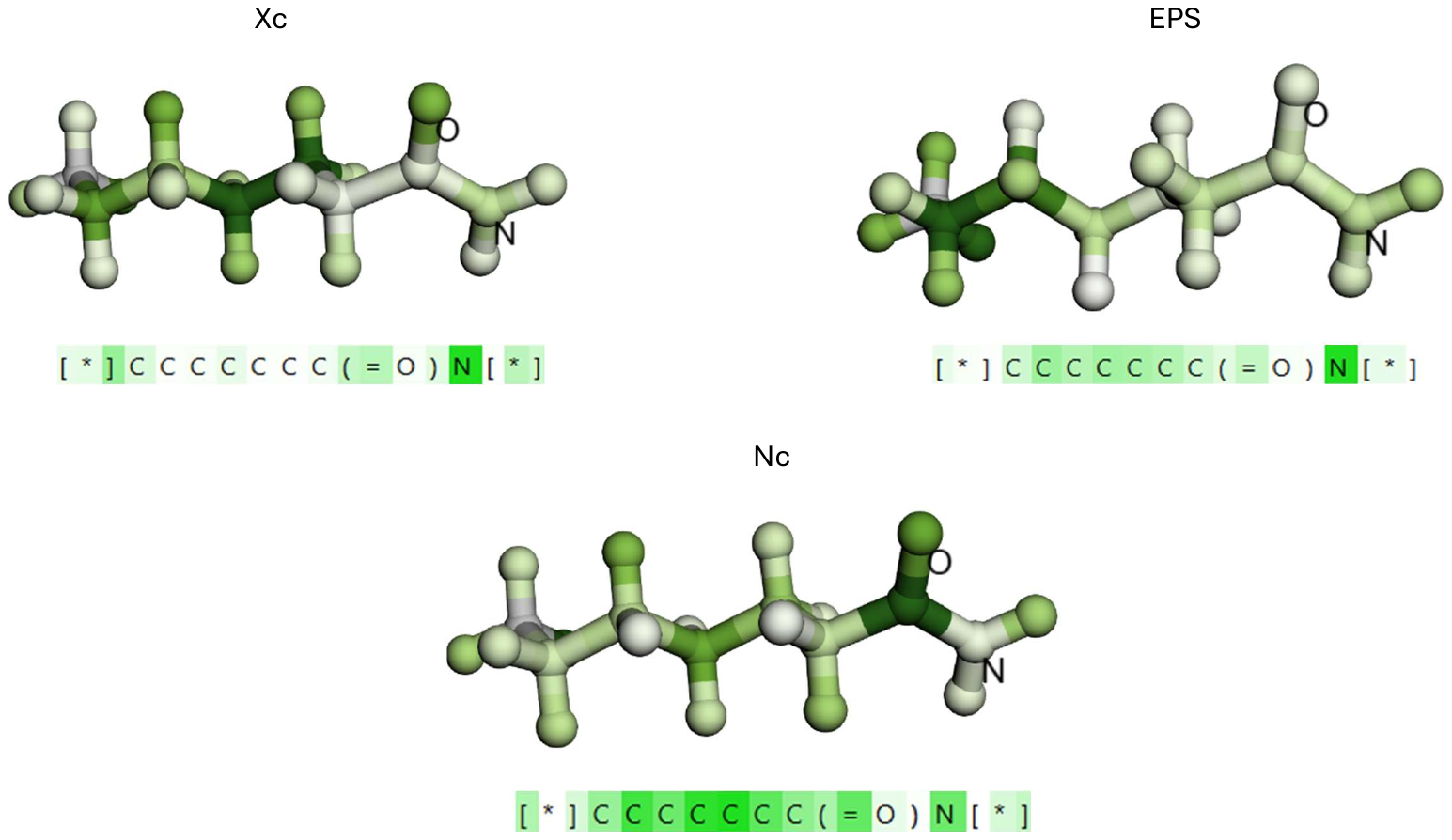}
  \caption{A plot of importance-score of the input structure in the case of [*]CCCCCCC(=O)N[*].
  Only oxygen and nitrogen atoms are labeled, whereas carbon and hydrogen atoms are left unlabeled for clarity. The plotted importance-score is normalized by the maximum value of the original importance-score.}
  \label{fig:XAI-1}
\end{figure}

Interpretability is a critical aspect of machine learning in chemistry, where understanding the structural features underlying model predictions can provide insights into the learned representations. In this section, we investigate the types of structural information utilized by the ML model to assess whether its predictions are associated with chemically meaningful features. To this end, we compute feature attribution scores with respect to the input data using Captum \cite{kokhlikyan2020captum}, which estimates feature importance based on gradient information\footnote{Gradient information was computed with respect to the embedding layer of the input tokens in the SMILES encoder and the atom-type node embedding layer of the MACE-MP model.}. By analyzing feature attributions across the different modalities, we examine whether the model consistently identifies chemically relevant functional groups and structural motifs in agreement with established chemical knowledge. Because many polymer properties are strongly influenced by specific functional groups, a chemically meaningful representation is expected to assign elevated attribution scores to atoms and substructures associated with these groups. This analysis provides additional evidence for the physical relevance of the learned representations and highlights the potential of multimodal approaches for interpretable polymer informatics.

As discussed above, molecular properties often exhibit strong dependencies on specific functional groups. In general, these relationships arise from the combined effects of electronic structure, such as donor and acceptor strength, dipole moment, polarizability, and three-dimensional chain conformation~\cite{darling2008isolating,ansari2018synthesis,wedler2020role,velasco2022assessment,mamand2025green,ma18010198,WU2025128546,aziz2025poly}. A summary of potentially relevant functional groups and their expected relationships with the target properties is provided in \autoref{tab:functional_groups_properties}. In the following analysis, we examine whether the ML model identifies these chemically relevant structural features directly from the data.

\autoref{fig:XAI} and \autoref{fig:XAI-1} present the importance scores for representative monomers with the following p-SMILES:
\begin{equation}
   \mathrm{[*]CCC(=O)O[*] \quad and \quad [*]CCCCCCC(=O)N[*]}.
\end{equation}
These p-SMILES correspond to structures containing several functional groups, including amine (NH$_2$), carbonyl (-C=O-), hydroxyl (-OH), and alkyl (-R, e.g., -CH$_3$).

We first consider the monomer [*]CCC(=O)O[*], shown in \autoref{fig:XAI}. In this structure, the ester linkage (-C(=)-O-) is located on the right-hand side, while the Alkyl groups (-C$_n$H$_{2n+1}$) appear on the left-hand side. Overall, both the SMILES-based and 3D models assign relatively high importance to these functional groups, though it also depends on the target properties. The SMILES model distributes attention across the entire p-SMILES string but exhibits enhanced attribution to tokens such as “(=)” and “O”, indicating that it more or less identifies chemically relevant substructures. In contrast, the 3D model yields more localized importance patterns, with particularly high scores in regions surrounding oxygen atoms, again consistent with chemically meaningful features. 

Next, we examine the monomer [*]CCCCCCC(=O)N[*], shown in \autoref{fig:XAI-1}. In this case, the carbonyl (-C=O) and amide linkage (-NH-) groups are located on the right-hand side of the structure. A similar trend is observed: both the SMILES-based and 3D models assign relatively high attribution scores to these functional groups, though depending on the target property. These results suggest that the model predictions are influenced, at least in part, by chemically meaningful structural motifs rather than being determined solely by global molecular features
\footnote{Although \autoref{tab:functional_groups_properties} summarizes whether individual functional groups tend to increase or decrease specific properties, the current Captum-based attribution analysis does not provide information about the directionality of these effects.}.

\begin{table}[htbp]
\begin{center}
\resizebox{\textwidth}{!}{
\begin{tabular}{lll}
\hline
\textbf{Functional group} & \textbf{Structure} & Mechanism \\
\hline

Carbonyl & C=O & Strong electron withdrawing \cite{peet2007efficiency,ansari2018synthesis}, planar conjugation enhances $\pi$-overlap \cite{VU2018128} \\

Fluoro & C--F, CF$_3$ & High electronegativity \cite{peet2007efficiency,ansari2018synthesis}; bulky groups disrupt packing\cite{WU2025128546}\\

Ether & C--O--C & Breaks conjugation \cite{VU2018128}, adds dipole \cite{ma18010198}\\

Alkyl & --C$_n$H$_{2n+1}$ & increases torsion/free volume \cite{wedler2020role} \\

Hydroxyl & --OH & H-bonding increases packing \cite{aziz2025poly} but breaks conjugation\cite{VU2018128}\\

Amine & --NH$_2$ & Electron-donating \cite{peet2007efficiency,ansari2018synthesis} \\

Sulfonyl & --SO$_2$--  & Strong dipole\cite{ma18010198} + electron-withdrawing \cite{peet2007efficiency,ansari2018synthesis}\\

\hline
\end{tabular}}
\end{center}
\caption{
Summary of functional group that can effect on polymer electronic, structural, and optical properties. 
}
\label{tab:functional_groups_properties}
\end{table}
\begin{figure}
  \centering
  \includegraphics[trim=0cm 0cm 0cm 0cm, 
                   clip,width=0.45\textwidth]{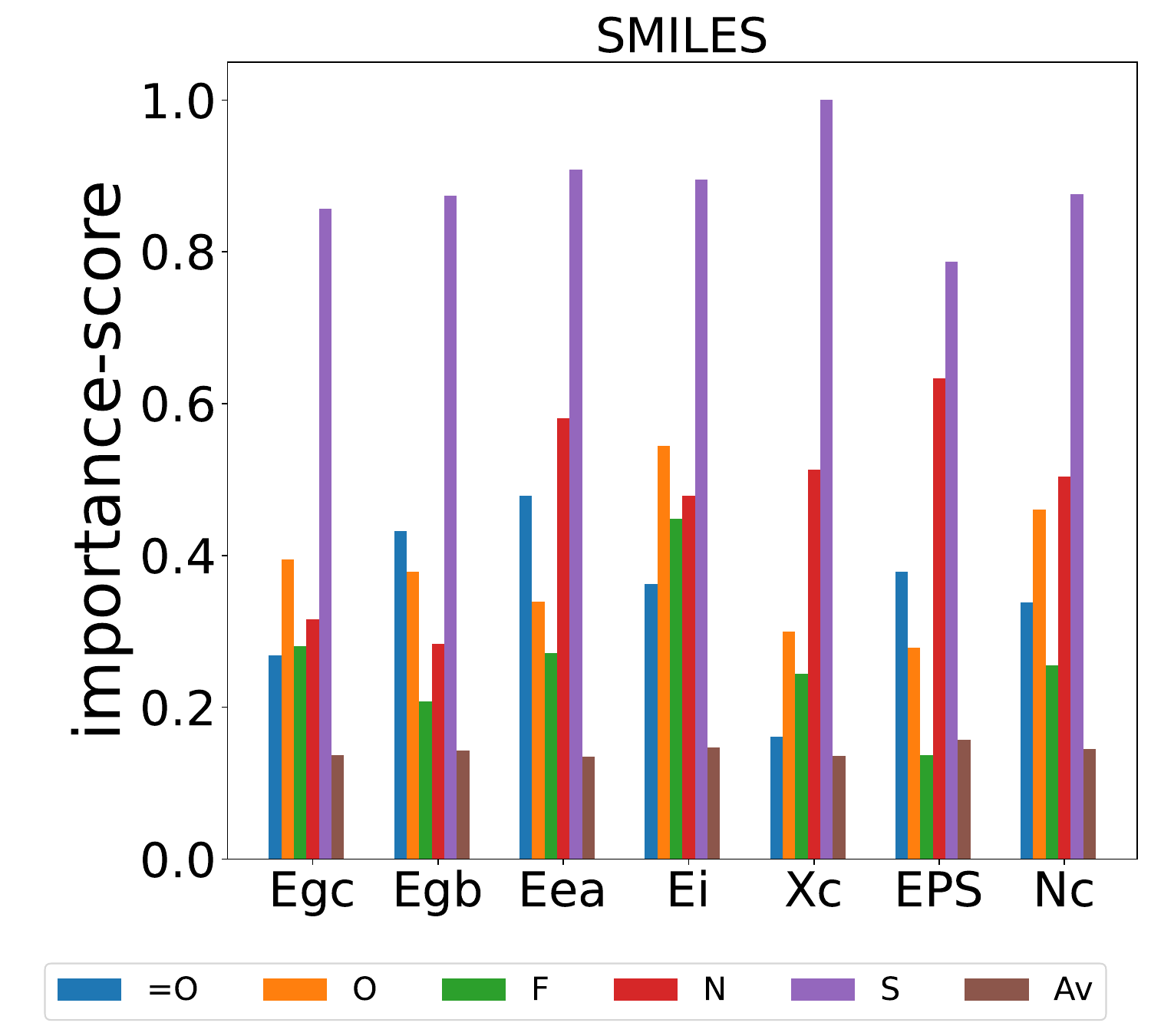}
  \includegraphics[trim=0cm 0cm 0cm 0cm, 
                   clip,width=0.45\textwidth]{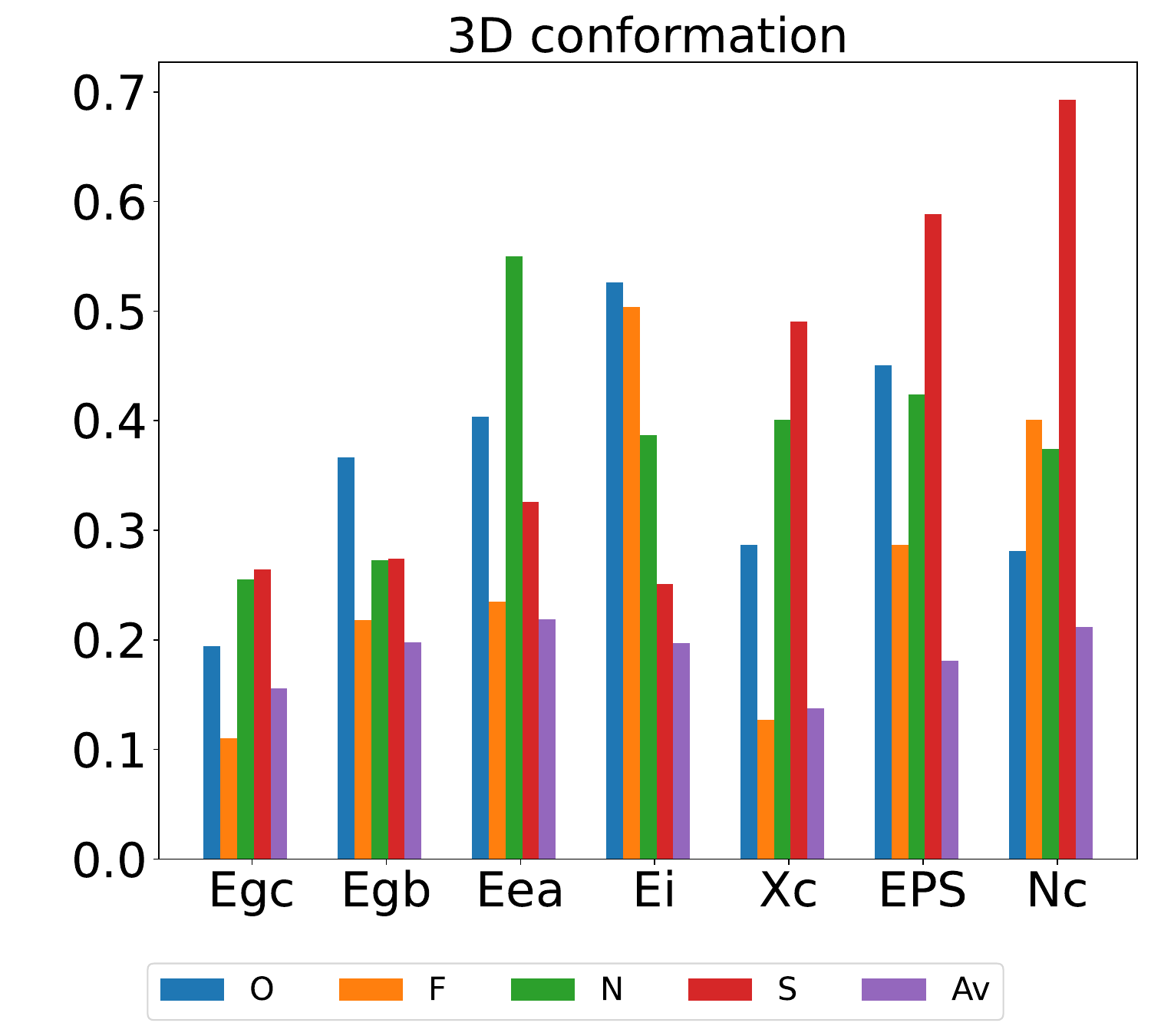}
  \caption{A statistics of importance-score of each dataset. "Av" denotes the average of the all the connectivity and atoms in the input data. The importance-score of each sample is normalized by the each sample's maximum value of the importance-score.} 
  \label{fig:XAI-bar}
\end{figure}

In \autoref{fig:XAI-bar}, we present the average importance scores of the atoms [=O, O, F, N, S]\footnote{Here, “O” denotes all occurrences of oxygen in the SMILES representation, including “=O”.} for the SMILES-based branch and [O, F, N, S] for the 3D conformation–based branch. Overall, the results indicate that the model attends preferentially to atoms associated with key functional groups. Notably, the SMILES importance score distribution suggests that the SMILES encoder emphasizes oxygen and sulfur related features while assigning comparatively less importance to fluorine. In contrast, the 3D encoder leverages a more diverse set of atomic features, with its attention patterns varying across datasets. 
Taken together, the attribution analyses indicate that both modality branches preferentially focus on chemically relevant heteroatoms and functional groups.
This observation motivates future investigations into incorporating explicit functional-group information as an inductive bias.

\section{Discussion}

In this work, we proposed a geometry-based fragmentation method that reduces reliance on predefined chemistry-specific fragmentation schemes, such as BRICS, while enabling flexible control over fragment size and the extraction of structural information at multiple length scales. By formulating molecular fragmentation as a radius-constrained optimization problem inspired by the k-center formulation, where the number of clusters is not fixed a priori but emerges from the specified radius, our approach can be applied to arbitrary molecular or material structures using purely geometric information, without requiring predefined chemical rules. Building on this framework, we developed a geometric fragment-based attention module and incorporated it into a multimodal model that combines p-SMILES representations with three-dimensional (3D) conformational information. The multimodal framework further integrates representative pretrained models for the respective modalities, allowing complementary sequence-based and geometry-aware representations to be exploited for polymer property prediction.

Although we use geometry-based fragments in the present model, the proposed fragmentation framework is not restricted to this particular strategy. Other fragmentation approaches, including chemistry-based methods such as BRICS, could potentially be incorporated into the same framework. Combining geometry-based and rule-based fragmentation may provide complementary inductive biases by integrating chemically predefined structural information with geometry-driven representations at multiple length scales. Exploring such hybrid fragmentation strategies represents an interesting direction for future work.

We systematically investigated the effects of fragment-based attention and multimodal information using a benchmark p-SMILES dataset for polymer property prediction. Incorporating 3D conformational information improves predictive performance for several target properties, with the multimodal model achieving an improvement over the baseline for several properties. However, the contribution of each modality is property-dependent. In particular, 3D structural information improves predictions for properties such as \egc\ and \xc \ of TransPolymer, whereas other properties exhibit greater sensitivity to the p-SMILES representation. The ablation analysis further reveals that the relative contributions of individual branches vary across datasets and target properties, with no single branch consistently outperforming the others. These results highlight the complementary nature of sequence-based and geometry-based representations and suggest that effective multimodal integration is important for capturing the diverse structural factors governing polymer properties.

To further investigate the structural information captured by the multimodal model, we performed explainable artificial intelligence (XAI) analysis using Captum, evaluating feature attributions with respect to the input representations. The results indicate that the model preferentially attends to chemically meaningful structural features. In both the SMILES-based and 3D representations, elevated attribution scores are observed for chemically relevant functional groups and heteroatoms, although the spatial patterns of attribution differ between the two modalities. The SMILES-based model distributes attribution more broadly across the input tokens, while the 3D model produces more localized attribution patterns around chemically relevant atomic regions. These observations suggest that the model learns representations that are associated with physically meaningful structural features and are broadly consistent with established chemical knowledge. However, the current attribution analysis does not resolve the directionality of these contributions; that is, it does not determine whether a given functional group increases or decreases a particular target property. Further work incorporating attribution methods capable of resolving feature-wise contribution directions could provide deeper insight into the learned structure--property relationships.

A potential limitation of the present approach concerns the accuracy and representativeness of the 3D conformations used as input. Although RDKit is widely used for molecular conformer generation in computational chemistry and machine learning research \cite{nguyen2024structure}, the conformations generated in our workflow are obtained using classical force-field-based optimization rather than quantum chemical methods such as density functional theory (DFT). Consequently, the generated structures may not accurately represent the energetically preferred conformations at the relevant level of theory. More importantly, flexible polymer structures can adopt a diverse ensemble of conformations, and multiple conformations may contribute to their observed macroscopic properties \cite{nguyen2024structure}. In the present work, we generate multiple conformers for each p-SMILES representation; however, only a single conformer is provided as the 3D input to the model at a given time, rather than simultaneously incorporating an ensemble of conformations as in \cite{nguyen2024structure}. Future work could therefore investigate higher-accuracy quantum-chemical conformers, explicit ensemble-based representations, or dynamically sampled conformations to better capture conformational variability and assess its impact on polymer property prediction. 

Overall, our results demonstrate the potential of geometry-based fragmentation as a chemistry-independent strategy for constructing multiscale structural representations and integrating them with complementary molecular modalities. The combination of geometry-based fragmentation, multimodal learning, and XAI provides a promising framework for developing polymer informatics models that are both predictive and physically interpretable. More broadly, because the proposed fragmentation method relies only on geometric information, it may be applicable beyond polymers and molecular systems to more general molecular and material structures, providing a flexible foundation for future multiscale and modality-aware machine learning approaches in computational chemistry and materials science.

\section{Data and code availability}

The training data are available from the TransPolymer repository at \url{https://github.com/ChangwenXu98/TransPolymer}. The PI1M dataset used for pretraining is publicly available at \url{https://github.com/RUIMINMA1996/PI1M}.
The source code and implementation details supporting the findings of this study are available from the corresponding author upon reasonable request. 


\section*{Acknowledgements}

Mathias Niepert acknowledges support from the Deutsche Forschungsgemeinschaft (DFG, German Research Foundation) under Germany’s Excellence Strategy (EXC 2075 – 390740016) and through grant 569019417. He also acknowledges support from the Stuttgart Center for Simulation Science (SimTech). 

\section*{Supporting information}

The following files are included in the ACS Omega journal version of our manuscript.
\begin{itemize}
  \item Supplemental\_Information.pdf: A more detailed model structure description, data preprocess, and training Setup; Additional details of geometry-based molecular fragmentation, including step-by-step example; Ablation studies of benchmark results; Functional-group distribution in fragments.

\end{itemize}

\bibliography{references}

@article{xu2023transpolymer,
  title={TransPolymer: a Transformer-based language model for polymer property predictions},
  author={Xu, Changwen and Wang, Yuyang and Barati Farimani, Amir},
  journal={npj Computational Materials},
  volume={9},
  number={1},
  pages={64},
  year={2023},
  publisher={Nature Publishing Group UK London}
}

@article{liu2019roberta,
  title={Roberta: A robustly optimized bert pretraining approach},
  author={Liu, Yinhan and Ott, Myle and Goyal, Naman and Du, Jingfei and Joshi, Mandar and Chen, Danqi and Levy, Omer and Lewis, Mike and Zettlemoyer, Luke and Stoyanov, Veselin},
  journal={arXiv preprint arXiv:1907.11692},
  year={2019}
}

@inproceedings{nguyen2024structure,
  title={Structure-Aware E (3)-Invariant Molecular Conformer Aggregation Networks},
  author={Nguyen, Duy Minh Ho and Lukashina, Nina and Nguyen, Tai and Le, An Thai and Nguyen, Trungtin and Ho, Nhat and Peters, Jan and Sonntag, Daniel and Zaverkin, Viktor and Niepert, Mathias},
  booktitle={International Conference on Machine Learning},
  pages={37736--37760},
  year={2024},
  organization={PMLR}
}

@article{kuenneth2023polybert,
  title={polyBERT: a chemical language model to enable fully machine-driven ultrafast polymer informatics},
  author={Kuenneth, Christopher and Ramprasad, Rampi},
  journal={Nature communications},
  volume={14},
  number={1},
  pages={4099},
  year={2023},
  publisher={Nature Publishing Group UK London}
}

@article{he2020deberta,
  title={Deberta: Decoding-enhanced bert with disentangled attention},
  author={He, Pengcheng and Liu, Xiaodong and Gao, Jianfeng and Chen, Weizhu},
  journal={arXiv preprint arXiv:2006.03654},
  year={2020}
}

@article{batatia2022mace,
  title={MACE: Higher order equivariant message passing neural networks for fast and accurate force fields},
  author={Batatia, Ilyes and Kovacs, David P and Simm, Gregor and Ortner, Christoph and Cs{\'a}nyi, G{\'a}bor},
  journal={Advances in neural information processing systems},
  volume={35},
  pages={11423--11436},
  year={2022}
}

@article{batatia2023foundation,
    author = {Batatia, Ilyes and Benner, Philipp and Chiang, Yuan and Elena, Alin M. and Kovács, Dávid P. and Riebesell, Janosh and Advincula, Xavier R. and Asta, Mark and Avaylon, Matthew and Baldwin, William J. and Berger, Fabian and Bernstein, Noam and Bhowmik, Arghya and Bigi, Filippo and Blau, Samuel M. and Cărare, Vlad and Ceriotti, Michele and Chong, Sanggyu and Darby, James P. and De, Sandip and Della Pia, Flaviano and Deringer, Volker L. and Elijošius, Rokas and El-Machachi, Zakariya and Fako, Edvin and Falcioni, Fabio and Ferrari, Andrea C. and Gardner, John L. A. and Gawkowski, Mikołaj J. and Genreith-Schriever, Annalena and George, Janine and Goodall, Rhys E. A. and Grandel, Jonas and Grey, Clare P. and Grigorev, Petr and Han, Shuang and Handley, Will and Heenen, Hendrik H. and Hermansson, Kersti and Ho, Cheuk Hin and Hofmann, Stephan and Holm, Christian and Jaafar, Jad and Jakob, Konstantin S. and Jung, Hyunwook and Kapil, Venkat and Kaplan, Aaron D. and Karimitari, Nima and Kermode, James R. and Kourtis, Panagiotis and Kroupa, Namu and Kullgren, Jolla and Kuner, Matthew C. and Kuryla, Domantas and Liepuoniute, Guoda and Lin, Chen and Margraf, Johannes T. and Magdău, Ioan-Bogdan and Michaelides, Angelos and Moore, J. Harry and Naik, Aakash A. and Niblett, Samuel P. and Norwood, Sam Walton and O’Neill, Niamh and Ortner, Christoph and Persson, Kristin A. and Reuter, Karsten and Rosen, Andrew S. and Rosset, Louise A. M. and Schaaf, Lars L. and Schran, Christoph and Shi, Benjamin X. and Sivonxay, Eric and Stenczel, Tamás K. and Sutton, Christopher and Svahn, Viktor and Swinburne, Thomas D. and Tilly, Jules and van der Oord, Cas and Vargas, Santiago and Varga-Umbrich, Eszter and Vegge, Tejs and Vondrák, Martin and Wang, Yangshuai and Witt, William C. and Wolf, Thomas and Zills, Fabian and Csányi, Gábor},
    title = {A foundation model for atomistic materials chemistry},
    journal = {The Journal of Chemical Physics},
    volume = {163},
    number = {18},
    pages = {184110},
    year = {2025},
    month = {11},
    issn = {0021-9606},
    doi = {10.1063/5.0297006},
    url = {https://doi.org/10.1063/5.0297006},
    eprint = {https://pubs.aip.org/aip/jcp/article-pdf/doi/10.1063/5.0297006/20801246/184110_1_5.0297006.pdf},
}

@article{kim2018polymer,
  title={Polymer genome: a data-powered polymer informatics platform for property predictions},
  author={Kim, Chiho and Chandrasekaran, Anand and Huan, Tran Doan and Das, Deya and Ramprasad, Rampi},
  journal={The Journal of Physical Chemistry C},
  volume={122},
  number={31},
  pages={17575--17585},
  year={2018},
  publisher={ACS Publications}
}

@article{queen2023polymer,
  title={Polymer graph neural networks for multitask property learning},
  author={Queen, Owen and McCarver, Gavin A and Thatigotla, Saitheeraj and Abolins, Brendan P and Brown, Cameron L and Maroulas, Vasileios and Vogiatzis, Konstantinos D},
  journal={npj Computational Materials},
  volume={9},
  number={1},
  pages={90},
  year={2023},
  publisher={Nature Publishing Group UK London}
}

@article{zhang2021fragat,
  title={FraGAT: a fragment-oriented multi-scale graph attention model for molecular property prediction},
  author={Zhang, Ziqiao and Guan, Jihong and Zhou, Shuigeng},
  journal={Bioinformatics},
  volume={37},
  number={18},
  pages={2981--2987},
  year={2021},
  publisher={Oxford University Press}
}

@inproceedings{wollschlager2024expressivity,
  title={Expressivity and Generalization: Fragment-Biases for Molecular GNNs},
  author={Wollschl{\"a}ger, Tom and Kemper, Niklas and Hetzel, Leon and Sommer, Johanna and G{\"u}nnemann, Stephan},
  booktitle={International Conference on Machine Learning},
  pages={53113--53139},
  year={2024},
  organization={PMLR}
}

@article{probst2018probabilistic,
  title={A probabilistic molecular fingerprint for big data settings},
  author={Probst, Daniel and Reymond, Jean-Louis},
  journal={Journal of cheminformatics},
  volume={10},
  number={1},
  pages={66},
  year={2018},
  publisher={Springer}
}

@article{rogers2010extended,
  title={Extended-connectivity fingerprints},
  author={Rogers, David and Hahn, Mathew},
  journal={Journal of chemical information and modeling},
  volume={50},
  number={5},
  pages={742--754},
  year={2010},
  publisher={ACS Publications}
}

@book{engel2018chemoinformatics,
  title={Chemoinformatics: basic concepts and methods},
  author={Engel, Thomas and Gasteiger, Johann},
  year={2018},
  publisher={John Wiley \& Sons}
}

@book{todeschini2008handbook,
  title={Handbook of molecular descriptors},
  author={Todeschini, Roberto and Consonni, Viviana},
  year={2008},
  publisher={John Wiley \& Sons}
}

@article{chen2021polymer,
  title={Polymer informatics: Current status and critical next steps},
  author={Chen, Lihua and Pilania, Ghanshyam and Batra, Rohit and Huan, Tran Doan and Kim, Chiho and Kuenneth, Christopher and Ramprasad, Rampi},
  journal={Materials Science and Engineering: R: Reports},
  volume={144},
  pages={100595},
  year={2021},
  publisher={Elsevier}
}

@article{capecchi2020one,
  title={One molecular fingerprint to rule them all: drugs, biomolecules, and the metabolome},
  author={Capecchi, Alice and Probst, Daniel and Reymond, Jean-Louis},
  journal={Journal of cheminformatics},
  volume={12},
  number={1},
  pages={43},
  year={2020},
  publisher={Springer}
}

@article{weininger1988smiles,
  title={SMILES, a chemical language and information system. 1. Introduction to methodology and encoding rules},
  author={Weininger, David},
  journal={Journal of chemical information and computer sciences},
  volume={28},
  number={1},
  pages={31--36},
  year={1988},
  publisher={ACS Publications}
}

@article{degen2008art,
  title={On the art of compiling and using'drug-like'chemical fragment spaces},
  author={Degen, Jorg and Wegscheid-Gerlach, Christof and Zaliani, Andrea and Rarey, Matthias},
  journal={ChemMedChem},
  volume={3},
  number={10},
  pages={1503},
  year={2008}
}

@article{ma2020pi1m,
  title={PI1M: a benchmark database for polymer informatics},
  author={Ma, Ruimin and Luo, Tengfei},
  journal={Journal of Chemical Information and Modeling},
  volume={60},
  number={10},
  pages={4684--4690},
  year={2020},
  publisher={ACS Publications}
}

@article{kokhlikyan2020captum,
  title={Captum: A unified and generic model interpretability library for pytorch},
  author={Kokhlikyan, Narine and Miglani, Vivek and Martin, Miguel and Wang, Edward and Alsallakh, Bilal and Reynolds, Jonathan and Melnikov, Alexander and Kliushkina, Natalia and Araya, Carlos and Yan, Siqi and others},
  journal={arXiv preprint arXiv:2009.07896},
  year={2020}
}

@article{VU2018128,
title = {Conjugated polymers: A systematic investigation of their electronic and geometric properties using density functional theory and semi-empirical methods},
journal = {Synthetic Metals},
volume = {246},
pages = {128-136},
year = {2018},
issn = {0379-6779},
doi = {https://doi.org/10.1016/j.synthmet.2018.10.007},
url = {https://www.sciencedirect.com/science/article/pii/S0379677918302807},
author = {Khanh B. Vu and Van V. Vu and Huong Phung {Thi Thu} and Ha N. Giang and Nguyen Minh Tam and Son Tung Ngo},
}

@article{mamand2025green,
  title={Green approach to synthesis polymer composites based on chitosan with desired linear and non-linear optical characteristics},
  author={Mamand, Dyari M and Hussen, Sarkawt A and Aziz, Shujahadeen B},
  journal={Scientific Reports},
  volume={15},
  number={1},
  pages={3130},
  year={2025},
  publisher={Nature Publishing Group UK London}
}

@Article{ma18010198,
AUTHOR = {Mogbojuri, Gabriel and Abtahi, Shaghayegh and Hendeniya, Nayanathara and Chang, Boyce},
TITLE = {The Effects of Chain Conformation and Nanostructure on the Dielectric Properties of Polymers},
JOURNAL = {Materials},
VOLUME = {18},
YEAR = {2025},
NUMBER = {1},
ARTICLE-NUMBER = {198},
URL = {https://www.mdpi.com/1996-1944/18/1/198},
PubMedID = {39795843},
ISSN = {1996-1944},
DOI = {10.3390/ma18010198}
}

@article{WU2025128546,
title = {Dielectric performance of all-organic polymer composites via incorporation of fluorinated molecules},
journal = {Polymer},
volume = {331},
pages = {128546},
year = {2025},
issn = {0032-3861},
doi = {https://doi.org/10.1016/j.polymer.2025.128546},
url = {https://www.sciencedirect.com/science/article/pii/S0032386125005324},
author = {Xudong Wu and Ze Zhang and Qinghe Wu and Yachin Ivry and Daniel Q. Tan},
}

@article{wedler2020role,
  title={What is the role of planarity and torsional freedom for aggregation in a $\pi$-conjugated donor--acceptor model oligomer?},
  author={Wedler, Stefan and Bourdick, Axel and Athanasopoulos, Stavros and Gekle, Stephan and Panzer, Fabian and McDowell, Caitlin and Nguyen, Thuc-Quyen and Bazan, Guillermo C and K{\"o}hler, Anna},
  journal={Journal of Materials Chemistry C},
  volume={8},
  number={14},
  pages={4944--4955},
  year={2020},
  publisher={Royal Society of Chemistry}
}

@article{darling2008isolating,
  title={Isolating the effect of torsional defects on mobility and band gap in conjugated polymers},
  author={Darling, Seth B},
  journal={The Journal of Physical Chemistry B},
  volume={112},
  number={30},
  pages={8891--8895},
  year={2008},
  publisher={ACS Publications}
}

@article{aziz2025poly,
  title={Poly (3-hexylthiophene) as a versatile semiconducting polymer for cutting-edge bioelectronics},
  author={Aziz, Ilaria Abdel and Tullii, Gabriele and Antognazza, Maria Rosa and Criado-Gonzalez, Miryam},
  journal={Materials Horizons},
  volume={12},
  number={15},
  pages={5570--5593},
  year={2025},
  publisher={Royal Society of Chemistry}
}

@article{ansari2018synthesis,
  title={Synthesis and characterization of poly (3-hexylthiophene): improvement of regioregularity and energy band gap},
  author={Ansari, Muhammad Azhar and Mohiuddin, Shaikh and Kandemirli, Fatma and Malik, Muhammad Imran},
  journal={RSC advances},
  volume={8},
  number={15},
  pages={8319--8328},
  year={2018},
  publisher={Royal Society of Chemistry}
}

@article{velasco2022assessment,
  title={Assessment of the refractive index and extinction coefficient of graphene-poly (3-hexylthiophene) nanocomposites},
  author={Velasco Davoise, Lara and Pe{\~n}a Capilla, Rafael and D{\'\i}ez-Pascual, Ana M},
  journal={Polymers},
  volume={14},
  number={9},
  pages={1828},
  year={2022},
  publisher={MDPI}
}

@article{peet2007efficiency,
  title={Efficiency enhancement in low-bandgap polymer solar cells by processing with alkane dithiols},
  author={Peet, Jeffery and Kim, Jin Young and Coates, Nelson E and Ma, Wang Li and Moses, Daniel and Heeger, Alan J and Bazan, Guillermo C},
  journal={Nature materials},
  volume={6},
  number={7},
  pages={497--500},
  year={2007},
  publisher={Nature Publishing Group UK London}
}

@article{wolf2011facility,
  title={Facility location: concepts, models, algorithms and case studies. series: Contributions to management science},
  author={Wolf, Gert W},
  journal={International Journal of Geographical Information Science},
  volume={25},
  number={2},
  pages={331--333},
  year={2011},
  publisher={Taylor \& Francis, Inc. Bristol, PA, USA}
}

@article{GONZALEZ1985293,
title = {Clustering to minimize the maximum intercluster distance},
journal = {Theoretical Computer Science},
volume = {38},
pages = {293-306},
year = {1985},
issn = {0304-3975},
doi = {https://doi.org/10.1016/0304-3975(85)90224-5},
url = {https://www.sciencedirect.com/science/article/pii/0304397585902245},
author = {Teofilo F. Gonzalez},
}

@inproceedings{sener2018active,
  title={Active Learning for Convolutional Neural Networks: A Core-Set Approach},
  author={Sener, Ozan and Savarese, Silvio},
  booktitle={International Conference on Learning Representations},
  year={2018}
}

@article{zaverkin2022,
    author = {Zaverkin, Viktor and Holzmüller, David and Steinwart, Ingo and Kästner, Johannes},
    title = {Exploring chemical and conformational spaces by batch mode deep active learning},
    journal = {Digital Discovery},
    volume = {1},
    number = {5},
    pages = {605-620},
    year = {2022},
    month = {10},
    issn = {2635-098X},
    doi = {10.1039/d2dd00034b},
    url = {https://doi.org/10.1039/d2dd00034b},
    eprint = {https://pubs.rsc.org/dd/article-pdf/1/5/605/1833960/d2dd00034b.pdf},
}

@article{roberts1965basic,
  title={Basic principles of organic chemistry},
  author={Roberts, John D and Caserio, Marjorie C},
  year={1965}
}

@book{mcmurry2005organic,
  title={The organic chemistry of biological pathways},
  author={McMurry, John and Begley, Tadhg P},
  year={2005},
  publisher={Roberts and Company Publishers}
}

@article{van2008visualizing,
  title={Visualizing data using t-SNE.},
  author={Van der Maaten, Laurens and Hinton, Geoffrey},
  journal={Journal of machine learning research},
  volume={9},
  number={11},
  year={2008}
}

@article{Hjorth2017,
	author={Ask Hjorth Larsen and Jens J{\o}rgen Mortensen and Jakob Blomqvist and Ivano E Castelli and Rune Christensen and Marcin Du{\l}ak and Jesper Friis and Michael N Groves and Bj{\o}rk Hammer and Cory Hargus and Eric D Hermes and Paul C Jennings and Peter Bjerre Jensen and James Kermode and John R Kitchin and Esben Leonhard Kolsbjerg and Joseph Kubal and Kristen Kaasbjerg and Steen Lysgaard and J{\'o}n Bergmann Maronsson and Tristan Maxson and Thomas Olsen and Lars Pastewka and Andrew Peterson and Carsten Rostgaard and Jakob Schi{\o}tz and Ole Sch{\"u}tt and Mikkel Strange and Kristian S Thygesen and Tejs Vegge and Lasse Vilhelmsen and Michael Walter and Zhenhua Zeng and Karsten W Jacobsen},
	title={The atomic simulation environment—a Python library for working with atoms},
	journal={J. Phys. Condens. Matter},
	volume={29},
	pages={273002},
	year={2017}
}

\end{document}